\PassOptionsToPackage{table}{xcolor}
\documentclass[sn-vancouver-ay, iicol]{sn-jnl}

\usepackage{graphicx}%
\usepackage{multirow}%
\usepackage{amsmath,amssymb,amsfonts}%
\usepackage{amsthm}%
\usepackage{mathrsfs}%
\usepackage[title]{appendix}%
\usepackage{textcomp}%
\usepackage{manyfoot}%
\usepackage{booktabs}%
\usepackage{algorithm}%
\usepackage{algorithmicx}%
\usepackage{algpseudocode}%
\usepackage{listings}%

\usepackage{todonotes}
\usepackage{colortbl}
\usepackage{subcaption}
\usepackage{caption}
\usepackage{xspace}
\usepackage{mathtools}
\usepackage{pifont}
\usepackage{esvect}
\usepackage{color,soul}

\newcommand{\cmark}{\ding{51}} 
\newcommand{\xmark}{\ding{55}} 
\newcommand*{\mlp}{{\text{MLP}}}

\definecolor{salmon}{RGB}{250, 128, 114}
\definecolor{mistyrose}{RGB}{255, 228, 225}
\definecolor{codebg}{HTML}{EAEAF1}

\definecolor{revisioncolor}{RGB}{230, 126, 34}
\newcommand{\rev}[1]{#1}

\theoremstyle{thmstyleone}%
\theoremstyle{thmstyletwo}%

\theoremstyle{thmstylethree}%

\begin{document}

\title{NOOUGAT \raisebox{-0.3em}{\includegraphics[height=1.5em]{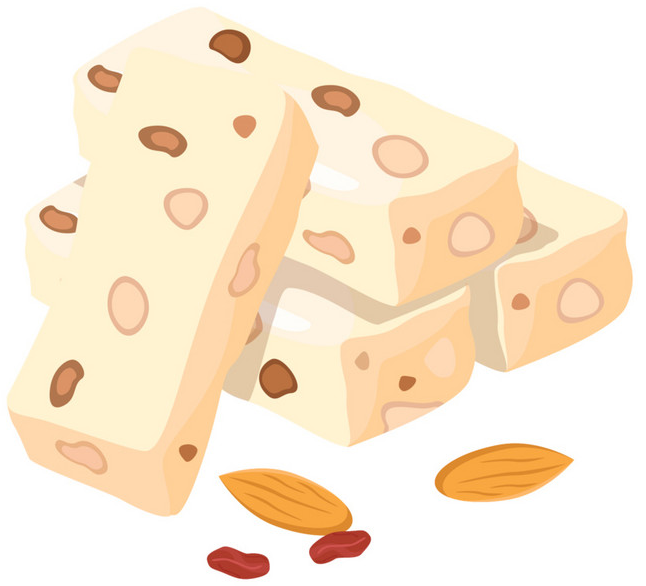}} Towards Unified Online and Offline Multi-Object Tracking}


\author*[1]{\fnm{Benjamin} \sur{Missaoui}}\email{bmissaoui@nvidia.com}
\author[1,2]{\fnm{Orcun} \sur{Cetintas}}
\author[1]{\fnm{Guillem} \sur{Brasó}}
\author[1]{\fnm{Tim} \sur{Meinhardt}}
\author[1]{\fnm{Laura} \sur{Leal-Taixé}}
\affil[1]{NVIDIA}
\affil[2]{Technical University of Munich}

\newcommand{\ben}[1]{{\leavevmode\color{orange}[Ben: #1]}}
\newcommand{\orc}[1]{{\leavevmode\color{teal}[Orcun: #1]}}
\newcommand{\gui}[1]{{\leavevmode\color{blue}[Guillem: #1]}}
\newcommand{\lau}[1]{{\leavevmode\color{magenta}[Khaleessi: #1]}}
\newcommand{\tim}[1]{{\leavevmode\color{red!70!black}[Tim: #1]}}
\newcommand{\old}[1]{{\leavevmode\color{gray}[Old text: #1]}}

\newcommand{\modelname}{NOOUGAT\@\xspace}
\newcommand{\layername}{Global GNN\@\xspace}
\newcommand{\PAR}[1]{{\noindent{\textbf{#1}}}} 
\newcommand{\eqdef}{\vcentcolon=}

\definecolor{revonecolor}{RGB}{205, 92, 92}   
\definecolor{revtwocolor}{RGB}{184, 134, 11}  
\definecolor{revthreecolor}{RGB}{128, 90, 150} 
\definecolor{revonebg}{RGB}{255, 235, 235}    
\definecolor{revtwobg}{RGB}{255, 248, 225}    
\definecolor{revthreebg}{RGB}{245, 240, 250}  
\newcommand{\Rone}{\textcolor{revonecolor}{\textbf{R1}}}
\newcommand{\Rtwo}{\textcolor{revtwocolor}{\textbf{R2}}}
\newcommand{\Rthree}{\textcolor{revthreecolor}{\textbf{R3}}}
\newcommand{\reviewerquotebox}[3]{%
  \par\noindent
  \begin{tikzpicture}
    \node[inner sep=5pt, fill=#1, draw=#2, rounded corners=3pt, anchor=north west] at (0,0) {%
      \begin{minipage}[t]{\dimexpr\linewidth-12pt\relax}%
        \fontsize{8}{9.5}\selectfont\sloppy #3\par%
      \end{minipage}%
    };
  \end{tikzpicture}
  \par\vspace{3pt}
}
\newcommand{\Ronequote}[1]{\reviewerquotebox{revonebg}{revonecolor!70!black}{\Rone: #1}}
\newcommand{\Rtwoquote}[1]{\reviewerquotebox{revtwobg}{revtwocolor!70!black}{\Rtwo: #1}}
\newcommand{\Rthreequote}[1]{\reviewerquotebox{revthreebg}{revthreecolor!70!black}{\Rthree: #1}}


\abstract{
The long-standing division between \textit{online} and \textit{offline} Multi-Object Tracking (MOT) has led to fragmented solutions that fail to address the flexible temporal requirements of real-world deployment scenarios. 
Current \textit{online} trackers rely on frame-by-frame hand-crafted association strategies and struggle with long-term occlusions, whereas \textit{offline} approaches can cover larger time gaps, but still rely on heuristic stitching for arbitrarily long sequences.
In this paper, we introduce NOOUGAT, the first tracker designed to operate with arbitrary temporal horizons. NOOUGAT leverages a unified Graph Neural Network (GNN) framework that processes non-overlapping subclips, and fuses them through a novel Autoregressive Long-term Tracking (ALT) layer. 
The subclip size controls the trade-off between latency and temporal context, enabling a wide range of deployment scenarios, from frame-by-frame to batch processing.
NOOUGAT achieves state-of-the-art performance across both tracking regimes, improving \textit{online} AssA by +2.3 on DanceTrack, +9.2 on SportsMOT, and +5.0 on MOT20, with even greater gains in \textit{offline} mode.
}


\maketitle

\begin{figure*}[!h]
  \centering
  \begin{subfigure}[b]{0.31\textwidth}
    \includegraphics[height=4cm]{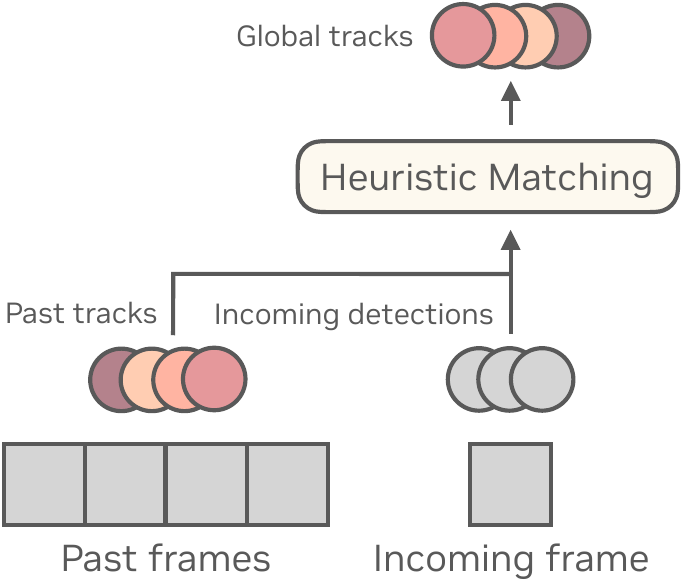}
    \caption{Online trackers}
    \label{fig:teaser_online}
  \end{subfigure}
  \hspace{0.02\textwidth}
  \begin{subfigure}[b]{0.31\textwidth}
    \includegraphics[height=4cm]{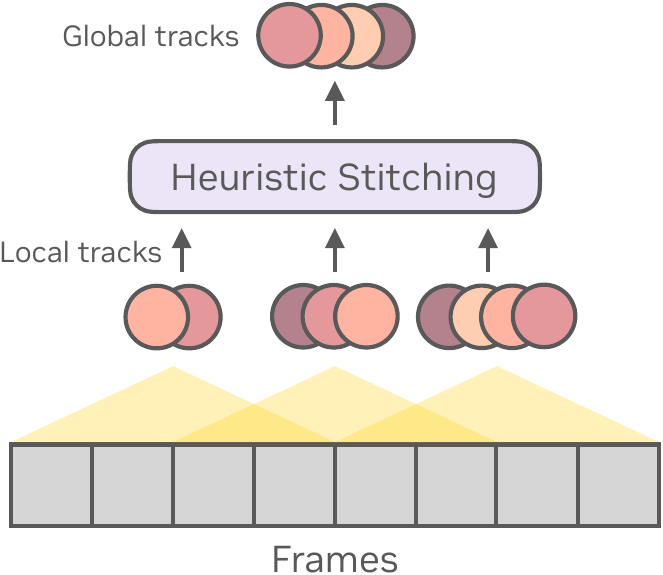}
    \caption{Offline trackers}
    \label{fig:teaser_offline}
  \end{subfigure}
  \hspace{0.02\textwidth}
  \begin{subfigure}[b]{0.31\textwidth}
    \includegraphics[height=4cm]{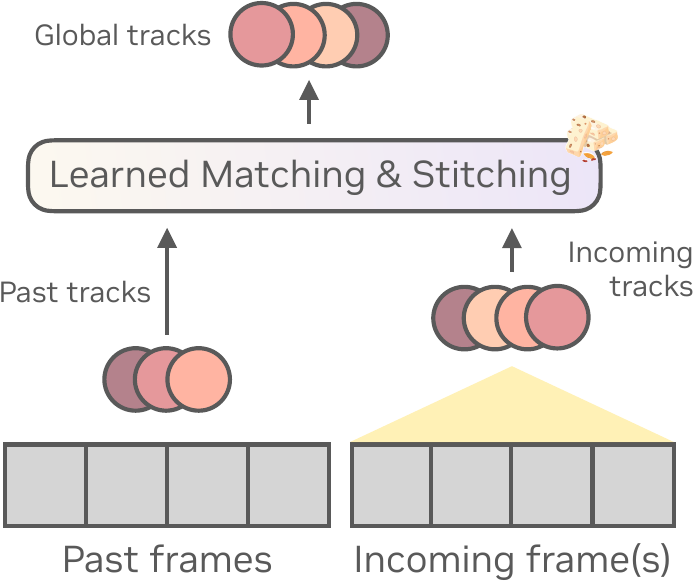}
    \caption{Our unified tracker}
    \label{fig:teaser_unified}
  \end{subfigure}
  \caption{(\ref{fig:teaser_online}) \textit{Online} trackers using heuristic matching. (\ref{fig:teaser_offline}) \textit{Offline} trackers using heuristics to stitch overlapping subclips. (\ref{fig:teaser_unified}) Our \modelname architecture eliminates the need for matching and stitching heuristics, and unifies \textit{online} and \textit{offline} in a single flexible framework.}
  \label{fig:teaser_comparison}
\end{figure*}

\section{Introduction} \label{sec:intro}

Multi-Object Tracking (MOT) aims at detecting objects and linking them across frames to form coherent trajectories. It is an essential task for many real-world systems, however, not all tracking applications have the same requirements. For instance, autonomous driving \cite{ding2024adatrackendtoendmulticamera3d, Luo_2021_ICCV, bdd100k} requires \textit{online} processing, where decisions must be made frame-by-frame using solely past information. In contrast, tasks such as dataset annotation \cite{spam2024eccv, 10.1007/978-3-642-15561-1_44} or post-event video analysis can be performed \textit{offline}, allowing access to future information to recover from occlusions and resolve identity switches. This inherent separation has driven the development of specialized models for each setting.

In \textit{online} tracking, while many works have attempted to design better motion \cite{lv2024diffmot, qin2023motiontracklearningrobustshortterm, cao2023observation, dendorfer2022quo, 6909848} and re-identification (ReID) \cite{wang2020highorderinformationmatterslearning, fu2020unsupervised, He_2021_ICCV, DBLP:conf/eccv/SomersAV24} models, the association module remains largely heuristic-driven \cite{Bewley2016_sort,Wojke2017simple}, often resorting to complex, hand-crafted multi-stage cascading strategies \cite{zhang2022bytetrack, cao2023observation, yang2024hybrid}. This leaves room for more principled, learned solutions. Additionally, although End-to-End (E2E) methods have recently gained traction, they tend to be resource-intensive and perform poorly in low-data regimes as they jointly learn object detection and tracking \cite{10.1007/978-3-031-19812-0_38, Zhang_2023, MeMOTR, MOTIP}.
Conversely, \textit{offline} methods have increasingly adopted learned approaches, with Graph Neural Networks (GNNs) showing strong performance by learning associations directly from the data \cite{braso_2020_CVPR, MPNTrackSeg, Cetintas_2023_CVPR, spam2024eccv, 10.1609/aaai.v38i3.27953}. However, these methods typically assume that the whole sequence can be processed in a single forward pass, an assumption that fails for arbitrarily long videos. In that case, heuristic stitching is still required to connect tracks from overlapping subclips. 

In this paper, we question the need for these separate models and heuristics, and we design a unified method that aims to satisfy the temporal requirements of any real-world deployment scenario. 
We introduce \textbf{\modelname}, a flexible \textbf{N}eural \textbf{O}nline and \textbf{O}ffline \textbf{U}nified \textbf{G}raph \textbf{A}rchitecture for \textbf{T}racking. We first partition the input sequence into non-overlapping subclips, and we generate local trajectories for each of them independently with a GNN hierarchy inspired by the offline tracker SUSHI~\cite{Cetintas_2023_CVPR}. 
These are then fused into global trajectories with our new Autoregressive Long-term Tracking (ALT) layer. The subclip size serves as a tunable hyperparameter that controls the processing stride: setting it to 1 enables frame-by-frame tracking - akin to \textit{online} trackers, while larger values allow for richer temporal reasoning for applications that permit it. 

For instance, with a 30 FPS input stream, a new frame arrives every 33ms. In latency-critical applications such as autonomous driving, where the perception stack must respond within 100ms~\cite{10.1145/3296957.3173191}, the tracker is required to produce outputs immediately as each frame arrives. In contrast, aerial vehicle tracking systems typically operate at 1–2 FPS~\cite{10.1007/978-3-031-73013-9_4, schmidt2012ais}, allowing the system to accumulate 15 to 30 frames before performing inference. These additional frames provide valuable temporal context and enable more robust handling of occlusions and more informed decisions. Finally, in fully offline tasks such as dataset annotation or sports analytics, latency is not a constraint, and the tracker can process hundreds of frames in a single pass to maximize accuracy.
NOOUGAT supports all these scenarios by design: it can operate with any number of incoming frames, with steadily increasing performance, as shown in Figure~\ref{fig:incom-frames-ablat}. This makes NOOUGAT a versatile solution capable of adapting to the temporal requirements of a wide range of real-world applications. Additional application-specific examples are discussed in Section~\ref{par:app-centric-track}.

Central to our design is our ALT layer, which is a fully learnable, data-driven association module. ALT is a GNN layer that builds a graph to connect historical trajectories with the incoming tracks. At inference time, it is applied autoregressively to track objects for arbitrarily long sequences. 
%
Unlike traditional methods that rely on hand-crafted matching and stitching heuristics, ALT learns both operations jointly, adapting to the most relevant cues across diverse tracking scenarios.
Moreover, as a result of our design and training recipe,
we observe that ALT is naturally able to handle long occlusions, a persistent challenge for current \textit{online} methods. Together with the ALT layer, NOOUGAT delivers state-of-the-art \textit{online} and \textit{offline} association performance, all in a unified and flexible framework that adapts to virtually any application. Our contributions are:


\begin{itemize} 
  %
  \item We propose NOOUGAT, the first tracking architecture that satisfies the delay requirements of various deployment scenarios. Through a unified formulation, we eliminate the need for matching and stitching heuristics commonly used in existing \textit{online} and \textit{offline} trackers, thus bridging the long-standing division in the field. 
\item We introduce the ALT layer, a fully learnable and data-driven GNN association module that dynamically uses the most relevant cues across various temporal contexts to perform robust association.
\end{itemize}

\begin{center}
    \centering
    \begin{figure}
        \centering
    \includegraphics[width=0.45\textwidth]{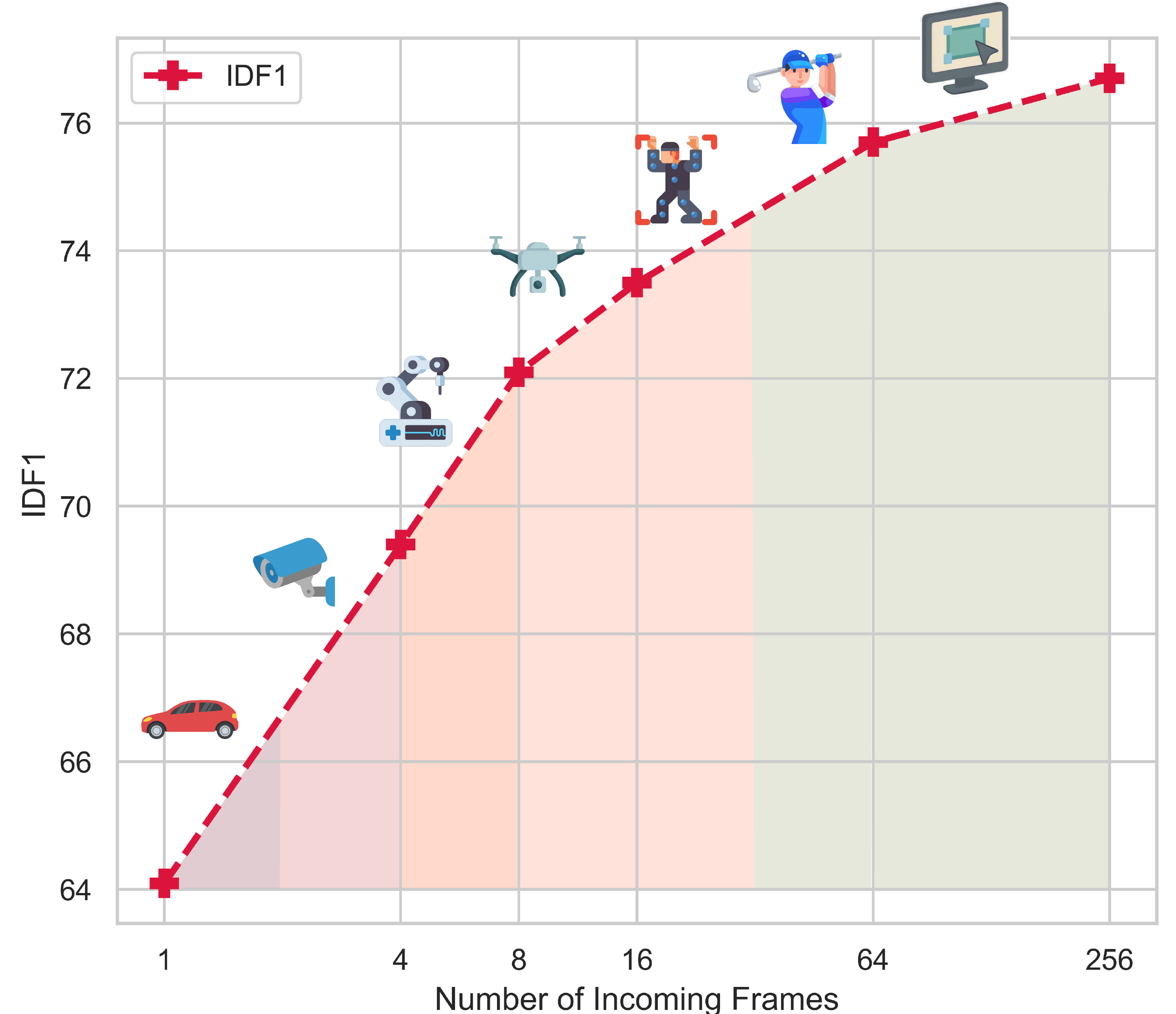}
        \caption{
        Impact of the number of incoming frames on DanceTrack val. Performance improves steadily from 1 to 256 frames, demonstrating NOOUGAT’s ability to scale with temporal context and adapt to diverse application requirements.}
        \label{fig:incom-frames-ablat}
    \end{figure}
\end{center}

\section{Related Work}
\label{sec:related-works}

\PAR{Online tracking.}
Modern \textit{online} trackers typically fall into two categories: SORT-based or End-to-End (E2E). SORT-based methods, built upon \cite{Bewley2016_sort}, follow the Tracking-by-Detection (TbD) paradigm, that decouples detection and tracking. 
While many works have proposed better motion \cite{cao2023observation, qin2023motiontracklearningrobustshortterm, lv2024diffmot, han2024ettrackenhancedtemporalmotion, 10.1145/3664647.3680944} and ReID \cite{wang2020highorderinformationmatterslearning,fu2020unsupervised,He_2021_ICCV,DBLP:conf/eccv/SomersAV24,10203280, DBLP:journals/corr/Leal-TaixeCS16} models, the core association step remains largely heuristic-driven, often implemented via single-stage \cite{Bewley2016_sort} or cascaded matching \cite{zhang2022bytetrack, aharon2022botsortrobustassociationsmultipedestrian, cao2023observation, seidenschwarz2023simplecuesleadstrong}. 
In contrast, E2E methods jointly learn detection and tracking within a unified architecture built upon DETR \cite{10.1007/978-3-030-58452-8_13, meinhardt2021trackformer, 10.1007/978-3-031-19812-0_38}, where track queries are used to detect and associate objects across frames. Despite their elegant design and recent progress, E2E approaches often struggle with long-term associations \cite{9880137} and require large-scale training data \cite{10.1007/978-3-031-19812-0_38, MeMOTR,Zhang_2023, MOTIP}, which is scarce in MOT. Other frameworks have explored alternative strategies, such as regressing object positions directly from detector outputs, as in Tracktor~\cite{tracktor_2019_ICCV}, or optimizing tracking performance through differentiable proxies of MOT metrics~\cite{xu2020train}.
In this work, our proposed Autoregressive Long-Term Tracking module (ALT) learns association in a data-driven manner, thus eliminating the need for the heuristic matching rules commonly used in SORT. Also, since we leverage off-the-shelf detectors, our model can focus on association, making it more lightweight and data-efficient than current E2E methods. 

\PAR{Offline tracking.}
Unlike frame-by-frame methods, \textit{offline} trackers aim to find globally optimal associations across multiple frames, enabling more robust and context-aware decisions. 
While some methods extend SORT with \textit{offline} heuristics \cite{du2023strongsort} or leverage transformers for global reasoning \cite{zhou2022global}, most \textit{offline} trackers adopt graph-based formulations. These approaches benefit from established optimization techniques such as network flows \cite{berclaz2011multiple, network_flows_tracking}, multi-cuts \cite{Tang_2017_CVPR}, minimum cliques \cite{zamir2012gmcp}, disjoint paths \cite{subgraph, lift, aplift}, and efficient solvers \cite{berclaz2011multiple, Butt2013}. Graphs naturally model detections as nodes and associations as edges, making them well-suited for handling occlusions and understanding object interactions. 
However, \textit{offline} trackers typically assume that the entire sequence can be processed in a single forward pass, an assumption that fails for longer sequences. Thus, to alleviate the computational constraints, these methods typically rely on heuristics, \textit{e.g.} linear programming, to stitch overlapping subclips into longer trajectories. 
This not only introduces hand-crafted design choices but also leads to redundant computations, as overlapping frames are processed multiple times.
In contrast, our ALT layer learns the stitching operation across non-overlapping subclips in a fully data-driven manner. 
This allows NOOUGAT to avoid both heuristic dependencies and redundant computations, resulting in a cleaner and more scalable architecture.

\PAR{Learning in graph-based tracking.}
 Early graph-based tracking methods relied on hand-crafted models or shallow learning techniques, such as conditional random fields \cite{9007500} and color-based similarity metrics \cite{color_tracking} to estimate pairwise association costs. 
More recent approaches have shifted towards deep learning, using convolutional networks to learn appearance-based affinities \cite{Leal-Taixe_2016_CVPR_Workshops, Son_2017_CVPR, Ristani_2018_CVPR}, and recurrent models to manage track states \cite{Sadeghian_2017_ICCV, tracking_rnn}. 
More recently, Graph Neural Networks (GNNs) have emerged as powerful tools for learning directly on graph structures \cite{braso_2020_CVPR, lpc, gmt, graph_nets_mot, gnn_3dmot, gsm, braso2022multi, Cetintas_2023_CVPR, 10.1609/aaai.v38i3.27953}. However, while these methods have shown strong performance in \textit{offline} settings, their use in \textit{online} tracking remains limited. 
%
%
%
Notable exceptions include \cite{10.1109/ICRA48506.2021.9561110} and \cite{9093347}, which use GNNs to refine node and edge embeddings but still rely on Hungarian Matching \cite{Kuhn1955Hungarian} for the final association. \cite{he2021gmtracker} introduces a Graph Matching layer, which performs association by solving a convex Quadratic Programming problem. Because their layer is differentiable, they can backpropagate to further refine the node embeddings. While their approach is elegant and learns data-driven associations, we find this extra step to be unnecessary.
Our architecture builds on the formulation of SUSHI~\cite{Cetintas_2023_CVPR}, which introduces a scalable GNN hierarchy to efficiently process long video clips. We generalize the framework beyond the original \textit{offline} formulation, in order to support not only \textit{online} settings, but also any specific delay requirements. Combined with our proposed ALT layer, NOOUGAT enables a general framework capable of processing arbitrarily long sequences in a fully learned fashion.

\section{Background}


\PAR{Tracking-by-Detection.} We follow the Tracking-by-Detection (TbD) paradigm. Given a set of object detections $\mathcal{O} =\{o_{i}\}_{i=1}^{n}$ for every frame of a video sequence, our task is to associate these detections into consistent trajectories. Each detection is defined by its bounding box coordinates, the corresponding image region, and its timestamp. The goal of the association stage is to construct a set of trajectories $\mathcal{T}$, where each trajectory $T_k\eqdef\{o_{k_i}\}_{i=1}^{n_{k}}$ consists of temporally ordered detections of the same object.

\PAR{Graph-based tracking. \label{par:graph-mot}}  We briefly review the standard graph-based formulation of the Multi-Object Tracking (MOT) problem \cite{zhang2008global}. The data association task is modeled as an undirected graph $G = (V, E)$, where each node $v \in V$ corresponds to an object detection, i.e., $V \coloneqq \mathcal{O}$. Edges $E \subset \{(o_i, o_j) \in V \times V \mid t_i \neq t_j\}$ represent potential associations between detections across different frames. A trajectory $T_k = \{o_{k_i}\}_{i=1}^{n_k}$, ordered by time such that $t_{k_i} < t_{k_{i+1}}$, forms a path in $G$ with edge set $E(T_k) \coloneqq \{(o_{k_1}, o_{k_2}), \dots, (o_{k_{n_k-1}}, o_{k_{n_k}})\}$. An edge $(u, v) \in E$ is labeled as correct if it belongs to a ground-truth trajectory, i.e., $y_{(u, v)} = 1$, and incorrect otherwise ($y_{(u, v)} = 0$). Given predicted edge scores $\{y^{\text{pred}}_{(u, v)}\}$, final trajectories are obtained with a flow solver or approximate heuristics. Overall, this approach frames MOT as a classification task over graph edges.

\PAR{Message-Passing GNNs. \label{par:mp-gnn}} Building on this graph formulation, MPNTrack~\cite{braso_2020_CVPR} proposes to use GNNs to learn a neural solver for MOT.
All nodes and edges are assigned embeddings that are initialized from association features. These embeddings are propagated through the graph and refined over $S$ message passing steps. 
Formally, given a graph $G = (V, E)$ with initial node embeddings $h^{(0)}_v \in \mathbb{R}^{d_V}$ and edge embeddings $h^{(0)}_{(u, w)} \in \mathbb{R}^{d_E}$, the embeddings are iteratively updated for $S$ steps. At the end, the final edge embeddings are classified via:
\begin{equation}
    y_{(u, v)}^{\text{pred}} = \mlp_{\text{class}}(h^{(S)}_{(u, v)}),
\end{equation}
and binarized via linear programming. 

\PAR{Hierarchical GNN. \label{par:gnn-hierarchy}} To scale GNNs to longer temporal horizons, SUSHI~\cite{Cetintas_2023_CVPR} proposes a hierarchical GNN architecture that recursively partitions the input clip. At the lowest level, nodes in the graph represent detections from adjacent frames, and a first GNN layer is used to associate them into short tracklets (e.g., frames 1–2, 3–4, ...). Higher levels treat each tracklet as a node, aggregating the geometry, motion, and appearance features of its individual detections. These tracklets are recursively merged into longer trajectories, up to the last level, which in the original paper covers a total of 512 frames. This hierarchical decomposition reduces graph complexity and edge density, offering a memory-efficient and scalable alternative to monolithic graph formulations. Additionally, GNN modules at different levels share the same weights, which is shown empirically to improve performance and convergence speed, while reducing the total number of learnable parameters. Our method builds on top of SUSHI, and uses a GNN hierarchy to efficiently generate tracklets for a set of non-overlapping subclips. However, unlike the original \textit{offline} formulation, we introduce a flexible design that accommodates \textit{online} tracking and variable delay requirements.
Furthermore, by replacing heuristic stitching with our learnable ALT module, our approach enables end-to-end data-driven association, scales to arbitrarily long sequences, and can bridge long-term occlusions.


\begin{center}
    \centering
    \begin{figure}[t]
        \centering
    \includegraphics[width=0.4\textwidth]{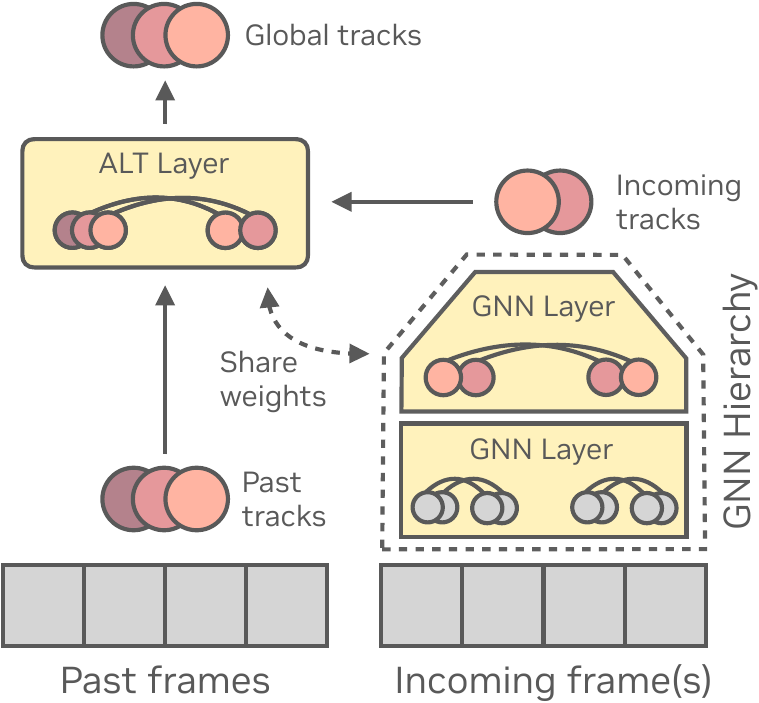}
        \caption{Overview of of \modelname. Our Global GNN module auroregressively connects past and incoming frames. It learns association across various temportal contexts in a data-driven manner, enabling both \textit{online} and \textit{offline} operation. }
        \label{fig:architecture}
    \end{figure}
\end{center}

\section{\modelname}

\textbf{NOOUGAT} is a unified and flexible framework for Multi-Object Tracking (MOT) that adapts to a wide range of application requirements. To process a sequence, NOOUGAT first partitions it into non-overlapping subclips of size $T$. We obtain tracklets for each subclip independently thanks to a GNN hierarchy. Then, our core component, the ALT layer, connects the tracklets from these subclips autoregressively to obtain global, sequence-length trajectories. 
At the first iteration, the tracklets from the first two subclips are merged. Then, ALT associates them with the tracklets from the next subclip, repeating this process until the entire sequence is covered. At any given iteration, we refer to the trajectories and frames from previously merged subclips as \textit{past tracks} and \textit{past frames} respectively, and to the tracklets and frames from the next subclip as \textit{incoming tracks} and \textit{incoming frames} respectively. 
A key design parameter in our architecture is the subclip size $T$, which determines the processing stride. With $T = 1$, at every iteration, NOOUGAT merges the past tracks with the detections in the incoming frame, thus behaving like an \textit{online} tracker.
Increasing $T$ enables more incoming frames to be processed jointly, thus providing richer temporal context and stronger performance. Thanks to this design, we are able to provide a model that maximizes performance for a wide range of applications. 
Our architecture is illustrated in Figure \ref{fig:architecture}. In the following sections, we give more details about our processing strategy and we go over the specifics of our ALT layer.

\subsection{Autoregressive Graph Tracking}
\label{sec:autoreg-tracking}
In this section, we give more details about our autoregressive pipeline, and how it differs from current graph methods.

\PAR{Limitations of sliding window inference. }
Given a video sequence $\mathcal{S}$ of $C$ frames, graph-based trackers typically divide $\mathcal{S}$ into $n_1$ overlapping subclips in a sliding window fashion. 
Each subclip has a fixed size of $T$ frames and the sliding window has stride $k < T$. This gives a set $X_1$ of subclips where each clip is processed independently by the graph solver:
\begin{equation}
    X_1 = \{s^1_{1\rightarrow T},s^2_{k\rightarrow T+k-1},  ..., s^{n_1}_{C-T+1\rightarrow C}\}
\label{eq:seq-partition}
\end{equation}

Having overlapping subclips is a necessary condition for this inference strategy, as it enables the trackers to stitch together the outputs of the different subclips - often via Linear Programming - to obtain sequence-length trajectories.
%
However, this strategy introduces a lot of redundant computation. 
SUSHI~\cite{Cetintas_2023_CVPR}, among other methods \cite{spam2024eccv, 10.1609/aaai.v38i3.27953}, sets the stride to $k=T/2$, meaning that the majority of the frames are processed twice. In \cite{braso_2020_CVPR, MPNTrackSeg}, clips of $T = 15$ frames are processed with stride $k=1$, meaning that each frame is processed up to 15 times. Figure \ref{fig:teaser_offline} illustrates these redundant computations. 


\PAR{Autoregressive inference. }
To avoid this unnecessary computational overhead, we adopt an autoregressive inference strategy.
We partition the sequence into $n_2$ non-overlapping subclips of size $T$, defined as in Equation \ref{eq:seq-autoreg-partition}:
\begin{equation}
    X_{2} = \{s^1_{1\rightarrow T}, s^2_{T+1\rightarrow 2T}, \dots, s^{n_2}_{C-T+1\rightarrow C}\}
\label{eq:seq-autoreg-partition}
\end{equation}
Each clip is processed independently by a GNN hierarchy, and merged autoregressively with our ALT module. 
Formally, we denote $\mathcal{T}^i_{alt}$ the set of trajectories produced by ALT up to subclip $s^i$, and $\mathcal{T}^i_{hicl}$ the set of trajectories produced by the hierarchical GNN for the subclip $s^i$\rev{, i.e., the local tracks within that subclip \textit{before} the autoregressive merging step}. We start by initializing:
\begin{equation}
    \mathcal{T}^1_{alt} := \mathcal{T}^1_{hicl} 
\end{equation}
Then, at each step $i = 2, ...,n_2$, we update our trajectories with:
\begin{equation}
    \mathcal{T}^i_{alt} := \text{ALT}(\mathcal{T}^{i-1}_{alt}, \mathcal{T}^i_{hicl})    
\end{equation}
where ALT$(\mathcal{T}_{past}, \mathcal{T}_{incom})$ denotes that we apply our ALT module to connect past and incoming tracklets.
After $n_2$ steps, we obtain our final set of tracks $\mathcal{T}^{n_2}_{ALT}$ which covers the entire input sequence.
As such, each frame is only processed once, which avoids redundant computations while remaining simple, flexible and scalable. 

\PAR{Graph connectivity mode.  \label{par:online-vs-offline} }
%
As explained in Section \ref{par:gnn-hierarchy}, in the original formulation proposed in SUSHI~\cite{Cetintas_2023_CVPR}, each node in the graph represents a trajectory and is connected to other nodes across both past and future frames. 
In our autoregressive notation, this can be formulated as ALT$(\mathcal{T}_{past}, \mathcal{T}_{incom})$ building a graph $G = (V, E)$, such that:
\begin{equation}
\begin{split}
V &=\{\mathcal{T}_{past} \cup \mathcal{T}_{incom}\} \\
E &= \{ (T_i, T_j) \in V \times V\}
\end{split}
\end{equation}
In contrast, real-time applications have strict inference requirements, as associations must be made using only past information. Furthermore, once a decision is made, it cannot be revised. This imposes a bipartite constraint on the graph, where all edges must connect past tracks with incoming ones\rev{; in this mode, nodes from the past connect \textit{only} to incoming nodes, preventing any modification of prior associations}. 
\noindent In that case, ALT$(\mathcal{T}_{past}, \mathcal{T}_{incom})$ builds a graph $G_{b} = (V, E_{b})$, such that:
\begin{equation}
\begin{split}
V &=\{\mathcal{T}_{past} \cup \mathcal{T}_{incom}\} \\
E_b &= \{ (T_i, T_j) \in \mathcal{T}_{past} \times \mathcal{T}_{incom}\}
\end{split}
\end{equation}
The ALT layer supports both the original graph formulation of SUSHI~\cite{Cetintas_2023_CVPR} and our bipartite extension. This flexibility allows NOOUGAT to accommodate both real-time and batch processing scenarios, and ensures fair comparisons with existing \textit{online} trackers, as discussed in Section~\ref{sec:results}.

\subsection{Unified Graph Architecture}
\label{sec:unif-arch-train}
In this section, we detail the node and edge features used by both the hierarchical GNN and the ALT module, and we describe our training procedure. These components collectively contribute to NOOUGAT’s ability to perform robust data association across varying temporal horizons and tracking scenarios within a unified architecture.

\PAR{Unified Node and Edge features}.
Since our architecture is fully learnable and data-driven, the model can adaptively select the most relevant association cues across different temporal contexts. To ensure consistency, we use a fixed set of input features across all sequences and datasets. Following SUSHI~\cite{Cetintas_2023_CVPR}, we use geometric, motion, and appearance cues as initial edge features. However, recent frame-by-frame trackers have demonstrated the added value of incorporating trajectory velocities \cite{cao2023observation, yang2024hybrid, lv2024diffmot} and detection confidences \cite{yang2024hybrid} to improve association. Thus, we propose to integrate both of these cues into our model.

\noindent Introduced in \cite{cao2023observation}, the Velocity Direction Consistency (VDC) is a cue designed specifically for frame-by-frame trackers. It captures the intuition that an incoming detection aligned with the direction of a past tracklet is more likely to be a correct match. VDC quantifies this by computing the angle between the tracklet’s velocity vector and the vector connecting it to a candidate detection. 
However, since the nodes in both the GNN hierarchy and ALT represent trajectories rather than individual detections, we extend the original VDC formulation to support this setting.

\noindent Formally, for an edge $(T_i,T_j)$, we compute $\vv{T_i},_{\text{fwrd}}$ the velocity vector of $T_i$ in the forward time direction and $\vv{T_j},_{\text{bwrd}}$ the velocity vector of $T_j$ in the backward time direction. \rev{The forward velocity $\vv{T_i},_{\text{fwrd}}$ is computed as the average displacement over the \textit{last} $\phi_1$ detections of a track, normalized to unit length; conversely, $\vv{T_j},_{\text{bwrd}}$ is computed from the \textit{first} $\phi_2$ detections. We set $\phi_1 = \phi_2 = 12$ across all configurations and datasets.} Then, the  $\text{VDC}$ is given by:
\begin{equation}
\text{VDC}(T_i,T_j) = \text{cos}^{-1} \left(\frac{
\vv{T_i},_{\text{fwrd}} \cdot (-\vv{T_j},_{\text{bwrd}})
}{
||\vv{T_i},_{\text{fwrd}} || \hspace{3pt}||\vv{T_j},_{\text{bwrd}}||
}\right)
\end{equation}
Once the VDC is computed, we add it to the initial set of edge features. Figure \ref{fig:ocm} illustrates the frame-by-frame VDC and our extension. We ablate this cue in the next section. 

\noindent For detection confidence, we follow the approach of \cite{spam2024eccv}, which adds bounding box dimensions and confidence scores as initial node embeddings. Although originally proposed for node classification - a task beyond the scope of this work, we find this representation to be an elegant and effective way to integrate detection confidence into the message-passing framework. 

\PAR{Training.}
\label{par:alt-training}
Graph-based trackers typically operate on fixed-size clips, ensuring relatively uniform tracklet lengths and occlusion patterns. In contrast, ALT must generalize across entire sequences, requiring a training process that reflects the diversity of temporal contexts encountered during inference.
To train our model, we begin by sampling a random subclip of \( T \) incoming frames, which is used to train the GNN hierarchy, following the setup in \cite{Cetintas_2023_CVPR}. Let \( t^* \) denote the first frame of this subclip. We then sample a random set of past frames \([k, p^*)\), for which we obtain ground-truth past tracks. \( p^* \) denotes the last past frame. We also drop random past tracks and detections as a data augmentation. ALT is trained to associate these ground-truth past tracklets with the output of the hierarchy.
To increase diversity, we randomize the start time $k$ and duration of the past tracklets, which we find improves generalization. Additionally, we introduce a small temporal gap between \( p^* \) and \( t^* \), which we refer to as the \textit{skip} parameter, that enables to simulate short occlusions and provides more realistic and challenging training examples.


\begin{center}
    \centering
    \begin{figure}[t]
        \centering
    \includegraphics[width=0.45\textwidth]{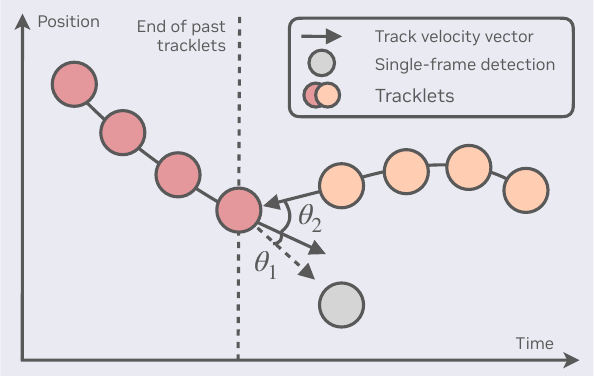}
        \caption{We extend the Velocity Direction Consistency (VDC) introduced in \cite{cao2023observation}. $\theta_1$ illustrates the original frame-by-frame VDC between a past track and an incoming detection, and $\theta_2$ is our extension to compute the similarity between past and incoming tracks.}
        \label{fig:ocm}
    \end{figure}
\end{center}

\section{Results}

\subsection{Datasets and Metrics}

We evaluate our approach on four diverse public benchmarks: DanceTrack~\cite{sun2022dance}, SportsMOT~\cite{cui2023sportsmot}, MOT17~\cite{dendorfer2020motchallengebenchmarksinglecameramultiple}, and MOT20~\cite{MOTChallenge20}, each presenting unique challenges in terms of motion dynamics, crowd density, and appearance variation.

\PAR{DanceTrack.} DanceTrack consists of 100 videos (106k frames) depicting group dance scenes. It is characterized by moderate crowding, high appearance similarity among individuals, and complex, non-linear motion conditions that pose significant challenges for association-based methods.

\PAR{SportsMOT.} SportsMOT focuses on high-speed sports footage, where rapid motion, frequent occlusions, and abrupt appearance changes are common. The dataset spans a variety of sports and emphasizes long-term identity preservation under dynamic conditions. It comprises 240 videos - including 150 sequences for the test set alone - with a total of 150k frames.

\PAR{MOT17.} MOT17 comprises both static and moving camera views across urban environments with varying pedestrian densities. It contains 14 sequences, 7 for training and 7 for testing, totaling 11k frames. In the public benchmark, detections are provided to isolate the association task, while the private setting allows the use of custom detections. We focus on the private setting.

\PAR{MOT20.} Designed to test tracking under extreme crowding, MOT20 includes sequences with over 200 pedestrians per frame. It contains 8 videos (13k frames), and more than 2 million objects. 

\PAR{Metrics.} We report standard MOT metrics to assess tracking performance: HOTA~\cite{hota}, AssA~\cite{hota}, IDF1~\cite{idf1}, MOTA~\cite{mota}. IDF1 captures identity preservation, MOTA emphasizes detection accuracy, and HOTA jointly evaluates detection, association, and localization. AssA is the HOTA submetric for association. Given our focus on association, AssA, IDF1 and HOTA serve as our primary metrics, and we also report MOTA for completeness.


\subsection{Implementation details}
\label{sec:implem-details}

\PAR{Detections and ReID.}
Following the protocol of ByteTrack~\cite{zhang2022bytetrack}, we use a YOLOX detector \cite{yolox2021} to generate detections for all four datasets, ensuring fair comparison with methods that also use the same detections. However, ByteTrack applies sequence-specific confidence thresholds on MOT17 and MOT20. To maintain consistency and generality, we adopt a fixed confidence threshold of 0.65 across all sequences and datasets, as done in \cite{Cetintas_2023_CVPR}. For MOT17 and MOT20, we apply offline linear interpolation to fill in missing detections, following prior work \cite{zhang2022bytetrack, aharon2022botsortrobustassociationsmultipedestrian, cao2023observation, yang2024hybrid}, since the ground truth includes annotations for fully occluded objects. No interpolation is applied on DanceTrack or SportsMOT. For ReID, we borrow the ResNet50-SBS model from FastReID \cite{he2020fastreid}, trained separately for each dataset, following the setup in \cite{aharon2022botsortrobustassociationsmultipedestrian, Wojke2017simple, yang2024hybrid, lv2024diffmot}. The models are then frozen during training.

\PAR{Processing stride.} We detail here the configurations that we refer to as NOOUGAT Online and NOOUGAT Offline in the results Section \ref{sec:results}. 
To ensure fair comparison with other trackers, we set the processing stride to $T=1$ in the \textit{online} setting, in order to perform frame-by-frame tracking. Also, we disable the correction of earlier associations, as described in Section \ref{par:online-vs-offline}. We adopt an Exponential Moving Average (EMA) to model the track's ReID features, as commonly done in the \textit{online} literature \cite{Wojke2017simple, yang2024hybrid, lv2024diffmot}. Note that with $T = 1$, the GNN hierarchy is not active, as ALT connects directly the past tracklets with the detections in the incoming frame.
For \textit{offline}, we set the processing stride to $T=256$, a value that we find to work well across datasets and enables to cover a wide range of occlusions and long-term associations. Since a node may be connected to both past and future frames, we do not adopt an EMA but instead we simply average the ReID features of all detections in a track.  

\PAR{Graph construction.} During both training and inference, our hierarchy construction follows \cite{Cetintas_2023_CVPR}. For the ALT layer, we connect each past track with its top 10 nearest neighbors within the incoming tracks, according to geometry, appearance and motion similarity. During training, we set the \textit{skip} parameter (see Section \ref{par:alt-training}), which controls the maximum gap between past and incoming tracks, to 8 frames.

\PAR{Training.} We train our \textit{online} tracker for 100k iterations with batch size 32 and our \textit{offline} version for 50k iterations with batch size 8. The ALT layer shares weights with the hierarchy, which we find empirically to improve convergence speed \rev{(roughly 40k iterations instead of 90k)}. We follow the progressive training scheme from \cite{Cetintas_2023_CVPR}, where we unfreeze each hierarchy level every 500 iterations. Once the hierarchy is fully unfrozen, we start training the ALT module. We use a focal loss \cite{8417976} with $\gamma=1$ and the Adam optimizer \cite{Kingma2014AdamAM}. We set the learning rate to $3\cdot 10^{-4}$ with a weight decay of $10^{-4}$.

\subsection{Benchmark results}
\label{sec:results}

\PAR{DanceTrack.} In Table~\ref{table:DanceTrack}, NOOUGAT outperforms all published work using ByteTrack's detections. In the \textit{online} setting, we report an improvement over the top heuristic model HybridSORT~\cite{yang2024hybrid} of 2.3 AssA and 3.2 IDF1. With the added temporal context, we gain an extra 3.8 AssA and 2.1 IDF1 in \textit{offline} mode. We significantly outperform the next best model CoNo-Link~\cite{10.1609/aaai.v38i3.27953}, which also uses GNNs.   
\begin{table}[!ht]
\centering
\resizebox{\columnwidth}{!}{%
\begin{tabular}{l|cccc}
\hline\noalign{\smallskip}
Tracker & HOTA & IDF1 & AssA  & MOTA\\
\noalign{\smallskip}
\hline
\textit{\color{gray}{Online trackers:}} & & &  \\
CenterTrack \cite{zhou2020tracking}                         & 41.8 & 35.7 & 22.6 & 86.8\\
FairMOT \cite{zhang2021fairmot}                             & 39.7 & 40.8 & 23.8 & 82.2\\
QDTrack \cite{qdtrack_conf}                             & 45.7 & 44.8 & 36.8 & 83.0\\
FineTrack \cite{10203280}                           & 52.7 & 59.8 & 38.5 & 89.9\\
\color{gray}{MOTRv2} \cite{Zhang_2023} & \color{gray}{69.9} & \color{gray}{71.7} & \color{gray}{59.0} & \color{gray}{91.9} \\
MeMOTR \cite{MeMOTR}                              & 68.5 & 71.2 & 58.4 & 89.9 \\
\rowcolor{salmon!15} SORT \cite{Bewley2016_sort}           & 47.9 & 50.8 & 31.2 & 91.8\\
\rowcolor{salmon!15} DeepSORT \cite{Wojke2017simple}       & 45.6 & 47.9 & 29.7 & 87.8\\
\rowcolor{salmon!15} ByteTrack \cite{zhang2022bytetrack}      & 47.3 & 52.5 & 31.4 & 89.5 \\
\rowcolor{salmon!15} GHOST \cite{seidenschwarz2023simplecuesleadstrong}          & 56.7 & 57.7 & 39.8 & 91.3 \\
\rowcolor{salmon!15} OC-SORT \cite{cao2023observation}        & 54.6 & 54.6 & 38.0 & 89.6\\
\rowcolor{salmon!15} Hybrid-SORT \cite{yang2024hybrid}    & 65.7 & 67.4 & 52.6 & 91.8\\
\rowcolor{salmon!15} DiffMOT \cite{lv2024diffmot}        & 61.3 & 63.0 & 47.2 & \textbf{92.8}\\
\rowcolor{salmon!15} \textbf{NOOUGAT} (ours)   & \textbf{65.9} & \textbf{70.6} & \textbf{54.9} & 88.9\\
\hline
\noalign{\smallskip}
\textit{\color{gray}{Offline Trackers:}} & & & \\
                     GTR \cite{zhou2022global}                               & 48.0 & 50.3 & 31.9  & 84.7\\
\rowcolor{salmon!15} StrongSORT++ \cite{du2023strongsort}                      & 55.6 & 55.2 & 38.6  & \textbf{91.1} \\
\rowcolor{salmon!15} SUSHI \cite{Cetintas_2023_CVPR}                             & 63.3 & 63.4 & 50.1  & 88.7\\
\rowcolor{salmon!15} CoNo-Link \cite{10.1609/aaai.v38i3.27953}                         & 63.8 & 64.1 & 50.7  & 89.7\\
\rowcolor{salmon!15} \textbf{NOOUGAT} (ours)   & \textbf{68.4} & \textbf{72.7} & \textbf{58.7} & 88.9\\
\hline
\end{tabular}
}
\caption{Results on the DanceTrack test set. Methods in the red block share the same detections. Methods in {\color{gray}gray} use extra training data.}
\label{table:DanceTrack}
\end{table}
\begin{table}[!ht]
\centering
\resizebox{\columnwidth}{!}{%
\begin{tabular}{l|cccc}
\hline\noalign{\smallskip}
Tracker & HOTA & IDF1 & AssA  & MOTA\\
\noalign{\smallskip}
\hline
\noalign{\smallskip}

\textit{\color{gray}{Online Trackers:}} & & & \\
CenterTrack \cite{zhou2020tracking}                                      & 62.7 & 60.0	&  48.0  & 90.8 \\
MeMOTR \cite{MeMOTR}                                           & 70.0 & 71.4 &  59.1  & 91.5 \\
MotionTrack \cite{qin2023motiontracklearningrobustshortterm}                                      & 74.0 & 74.0 &  61.7  & 96.6 \\
Deep-EIoU \cite{huang2024iterative}                                        & 77.2 & 79.8 &  67.7 & 96.3 \\
\rowcolor{salmon!15} ByteTrack \cite{zhang2022bytetrack}                   & 64.1 & 71.4 &  52.3  & 95.9 \\
\rowcolor{salmon!15} MixSort-Byte \cite{cui2023sportsmot}                & 65.7 & 74.1 &  54.8  & 96.2 \\
\rowcolor{salmon!15} OC-SORT \cite{cao2023observation}                      & 73.7 & 74.0 &  61.5  & 96.5 \\
\rowcolor{salmon!15} MixSort-OC \cite{cui2023sportsmot}                  & 74.1 & 74.4 &  62.0  & 96.5 \\
\rowcolor{salmon!15} DiffMOT \cite{lv2024diffmot}                     & 76.2 & 76.1 &  65.1  & \textbf{97.1} \\
\rowcolor{salmon!15} \textbf{\modelname}  (ours)  & \textbf{81.0} & \textbf{85.3} & \textbf{74.3} & 96.0\\
\hline 
\textit{\color{gray}{Offline Trackers:}} & & & \\
GTR \cite{zhou2022global}                  & 54.5 & 55.8 & 45.9 & 67.9 \\
\rowcolor{salmon!15} \textbf{\modelname}  (ours)  & \textbf{85.6} & \textbf{92.3} & \textbf{83.0}  &  \textbf{95.9}\\

\hline
\end{tabular}
}
\caption{Results on the SportsMOT test set. Methods in the red block share the same detections.}
\label{table:SPORTSMOT}
\end{table}

\PAR{SportsMOT.} We report a remarkable improvement over the state-of-the-art in Table \ref{table:SPORTSMOT}, with +9.2 AssA and +9.2 IDF1 compared to DiffMOT. SportsMOT presents a significant challenge because of its highly non-linear motion scenarios, thus these results highlight the versatility of our model to learn the right cues for different datasets. Although no recent \textit{offline} model has been submitted to SportsMOT, our model achieves another 8.7 AssA and 7.0 IDF1 performance jump compared to its \textit{online} counterpart. This emphasizes our capability to deliver the best performance for any processing stride.

\PAR{MOT17.} Beyond the varying camera viewpoints and pedestrian densities, the main challenge in MOT17 comes from the size of the dataset. With 5.9k frames, it contains around 5 times less training data than SportsMOT, and 8 times less than DanceTrack. This causes heuristic trackers like OC-SORT ~\cite{cao2023observation} and HybridSORT~\cite{yang2024hybrid} to usually perform better than End-to-End models like MeMOTR~\cite{MeMOTR} and MOTRv2~\cite{Zhang_2023}, even when the latter use extra training data like CrowdHuman~\cite{shao2018crowdhumanbenchmarkdetectinghuman}. Since our model focuses only on association, it is very lightweight and thus provides strong performance, even in the low data regime. For instance, our \textit{online} model has only 27K trainable parameters, which is 3 orders of magnitude less than the 51M found in MeMOTR. Because of this, NOOUGAT achieves state-of-the-art performance, outperforming DiffMOT by 0.7 AssA and 0.7 IDF1, even though DiffMOT uses extra data to train its motion model. In \textit{offline} mode, we also outperform CoNo-Link in terms of AssA and IDF1. However, CoNo-Link's strong detection model gives it the edge in HOTA. 

\PAR{MOT20.} 
Finally, in MOT20's crowded scenes, we achieve strong association capabilities, with improvements of 7.0 AssA compared to DiffMOT and 1.8 IDF1 compared to HybridSORT. These gains underscore the generalization strength of our framework across challenging scenarios. For \textit{offline}, we observe a similar pattern as MOT17, with stronger association performance but weaker detection than CoNo-Link.

\noindent Overall, we observe consistent association improvements across datasets, which demonstrates the generality of NOOUGAT.

\begin{table}[!ht]
\centering
\resizebox{\columnwidth}{!}{%
\begin{tabular}{l|cccc}
\hline\noalign{\smallskip}
Tracker & HOTA & IDF1 & AssA  & MOTA\\
\noalign{\smallskip}
\hline
\textit{\color{gray}{Online Trackers:}}  & & & & \\
CenterTrack \cite{zhou2020tracking}                                                & 52.2 & 64.7 & 51.0 & 67.8 \\
QDTrack \cite{qdtrack_conf}                                                    & 53.9 & 66.3 & 52.7 & 68.7 \\
FairMOT \cite{zhang2021fairmot}                                                    & 59.3 & 72.3 & 58.0 & 73.7 \\
MotionTrack \cite{qin2023motiontracklearningrobustshortterm}                                                & 65.1 & 80.1 & 60.2 & 65.1 \\
TrackFormer \cite{meinhardt2021trackformer}                                                & 57.3 & 68.0 & 54.1 & 74.1 \\
\color{gray}{MOTRv2} \cite{Zhang_2023}                        & \color{gray}{62.0} & \color{gray}{75.0} & \color{gray}{60.6} & \color{gray}{78.6} \\
\color{gray}{MeMOTR} \cite{MeMOTR}                        & \color{gray}{58.8} & \color{gray}{71.5} & \color{gray}{58.4} & \color{gray}{72.8} \\
\rowcolor{salmon!15} ByteTrack \cite{zhang2022bytetrack}                             & 63.1 & 77.3 & 62.0 & 80.3 \\
\rowcolor{salmon!15} GHOST \cite{seidenschwarz2023simplecuesleadstrong}                                 & 62.8 & 77.1 &   -  & 78.7 \\
\rowcolor{salmon!15} OC-SORT \cite{cao2023observation}                               & 63.2 & 77.5 & 63.4 & 78.0 \\
\rowcolor{salmon!15} Hybrid-SORT \cite{yang2024hybrid}                           & 64.0 & 78.7 &   -  & 79.9 \\
\rowcolor{salmon!15} \color{gray}{DiffMOT} \cite{lv2024diffmot}  & \color{gray}{64.5} & \color{gray}{79.3} & \color{gray}{64.6} & \color{gray}{79.8} \\
\rowcolor{salmon!15} \textbf{\modelname}  (ours)            & \textbf{65.2}   & \textbf{80.0}    &   \textbf{65.3}  & \textbf{80.7}    \\
\hline
\noalign{\smallskip}
\textit{\color{gray}{Offline Trackers:}}  & & & \\
GTR \cite{zhou2022global} & 59.1 & 71.5 & 57.0 & 75.3 \\
\rowcolor{salmon!15} StrongSORT++ \cite{du2023strongsort}                          & 64.4 & 79.5 & 64.4 & 79.6 \\
\rowcolor{salmon!15} SUSHI \cite{Cetintas_2023_CVPR}                                 & 66.5 & 83.1 & 67.8 & 81.1 \\
\rowcolor{salmon!15} CoNo-Link \cite{10.1609/aaai.v38i3.27953}                             & \textbf{67.1} & 83.7 & 67.8 & \textbf{82.7} \\
\rowcolor{salmon!15} \textbf{\modelname} (ours)             & 66.9 & \textbf{83.9} & \textbf{68.5} & 80.7 \\
\hline
\end{tabular}
}
\caption{Results on the MOT17 test set with private detections. Methods in the red block share the same detections. Methods in {\color{gray}gray} use extra training data.}
\label{table:MOT17}
\end{table}

\begin{table}[!ht]
\centering
\resizebox{\columnwidth}{!}{%
\begin{tabular}{l|cccc}
\hline\noalign{\smallskip}
Tracker & HOTA & IDF1 & AssA  & MOTA\\
\noalign{\smallskip}
\hline
\textit{\color{gray}{Online Trackers:}} \\
FairMOT \cite{zhang2021fairmot}                              & 54.6 & 67.3 & 54.7 & 61.8 \\
TrackFormer \cite{meinhardt2021trackformer}                          & 54.7 & 65.7 & 53.0 & 68.6 \\
FineTrack \cite{10203280}                             & 63.6 & 79.0 & 64.8 & 77.9 \\
MotionTrack \cite{qin2023motiontracklearningrobustshortterm}                          & 62.8 & 76.5 & 56.8 & 78.0 \\
\color{gray}{MOTRv2} & \color{gray}{61.0} & \color{gray}{73.1} & \color{gray}{59.3} & \color{gray}{76.2} \\
\rowcolor{salmon!15} ByteTrack \cite{zhang2022bytetrack}       & 61.3 & 75.2 & 59.6 & \textbf{77.8}\\
\rowcolor{salmon!15} BoT-SORT \cite{aharon2022botsortrobustassociationsmultipedestrian}        & 63.3 & 77.5 & 62.9 & \textbf{77.8} \\
\rowcolor{salmon!15} GHOST \cite{seidenschwarz2023simplecuesleadstrong}           & 61.2 & 75.2 &  -  & 73.7 \\
\rowcolor{salmon!15} OC-SORT \cite{cao2023observation}         & 62.1 & 75.9 & 62.5 & 75.5 \\
\rowcolor{salmon!15} Hybrid-SORT \cite{yang2024hybrid}     & 63.9 & 78.4 &  -  & 76.7 \\
\rowcolor{salmon!15} \color{gray}{DiffMOT} \cite{lv2024diffmot} & \color{gray}{61.7} & \color{gray}{74.9} & \color{gray}{60.5} & \color{gray}{76.7} \\
\rowcolor{salmon!15} \textbf{\modelname} (ours)            & \textbf{64.6}  & \textbf{80.2}   & \textbf{67.5} & 74.7 \\

\hline
\noalign{\smallskip}
\textit{\color{gray}{Offline Trackers:}} \\
\rowcolor{salmon!15} StrongSORT++  \cite{du2023strongsort}                    & 62.6 & 77.0 & 64.0 & 73.8\\
\rowcolor{salmon!15} SUSHI \cite{Cetintas_2023_CVPR}                            & 64.3 & 79.8 & 67.5 & 74.3\\
\rowcolor{salmon!15} CoNo-Link \cite{10.1609/aaai.v38i3.27953}                        & 65.9 & 81.8 & 68.0 & \textbf{77.5}\\
\rowcolor{salmon!15} \textbf{\modelname} (ours)        & \textbf{66.1} & \textbf{83.0} & \textbf{70.7} & 74.5\\
\hline
\end{tabular}
}
\caption{Results on the MOT20 test set with the private detections. Methods in the red block share the same detections. Methods in {\color{gray}gray} use extra training data.}
\label{table:MOT20}
\end{table}

%






\subsection{Ablation studies}

\noindent \PAR{Long-Term Tracking. \label{par:retention-window}}
Recovering identities after long occlusions is a persistent limitation of heuristic-based trackers. In most SORT-based methods, objects are considered lost if they remain undetected for more than 30 frames (i.e., one second at 30 FPS) \cite{Bewley2016_sort, Wojke2017simple, cao2023observation, yang2024hybrid, lv2024diffmot}. We refer to this threshold as the \textit{retention window} \rev{ (also sometimes known as ``max age'' in prior work~\cite{Wojke2017simple, he2021gmtracker})}. To mitigate this, several methods introduce specialized heuristics, such as dedicated association stages \cite{cao2023observation, yang2024hybrid, lv2024diffmot}, to reconnect with long-lost targets.
However, our ALT layer learns to handle such cases naturally. To evaluate this, we train multiple \textit{online} models on SportsMOT using increasing retention windows. We select SportsMOT for this ablation due to its high number of long occlusions. Figure~\ref{fig:retention} illustrates the results: the top panel compares our model (orange) with an oracle (red) that performs perfect edge classification; the bottom panel shows the distribution of occlusion durations. The steady improvement beyond the 30-frame mark highlights ALT’s ability to learn robust long-term associations in a data-driven manner, without relying on explicit heuristics.

\begin{center}
    \centering
    \begin{figure}
        \centering
    \includegraphics[width=0.45\textwidth]{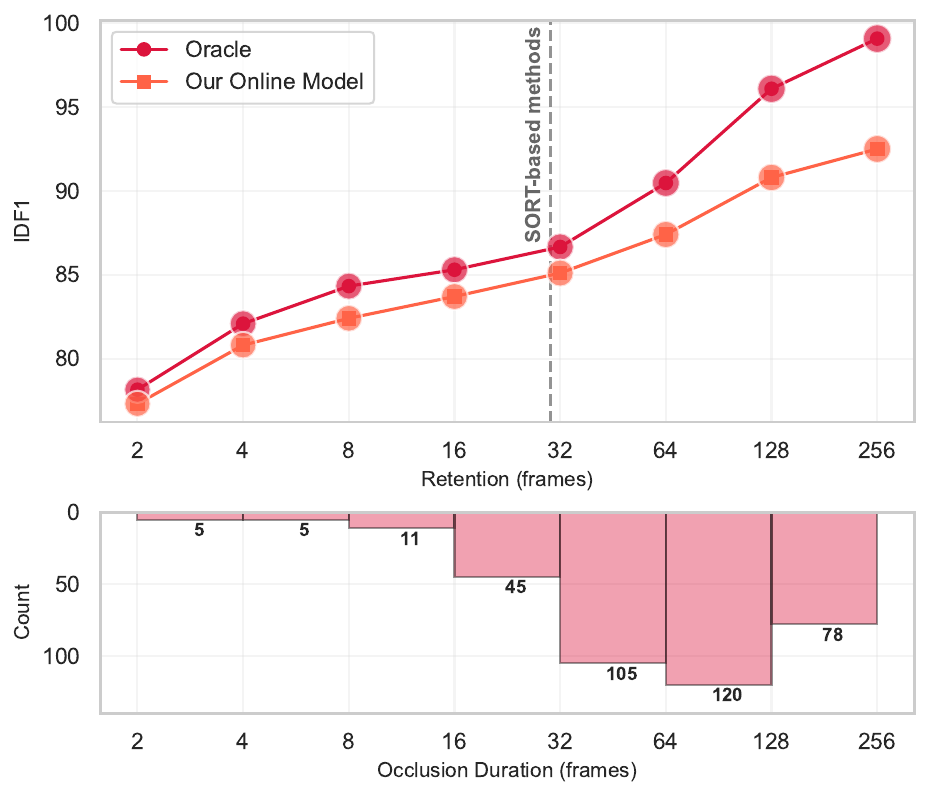}
        \caption{Oracle VS model performance on SportsMOT val set. SORT-based methods retain inactive tracks for very short timespans (usually 30 frames), and often rely on additional heuristics to recover lost tracks. Our model naturally learns to recover from long-term occlusions.}
        \label{fig:retention}
    \end{figure}
\end{center}

\PAR{Heuristic vs. Learnable Matching. }
\noindent To assess the effectiveness of our data-driven ALT association module, we replace it with the Hungarian algorithm~\cite{Kuhn1955Hungarian} and evaluate it \textit{online} on the DanceTrack validation set. All other parameters are kept constant, including the retention window of 256 frames.
The Hungarian matcher uses our three primary association cues: appearance, motion, and VDC. Each cue produces an individual cost matrix. We combine them with a weighted sum using coefficients of 4.0 for appearance, 1.0 for motion, and 0.2 for VDC, values that we borrow from HybridSORT~\cite{yang2024hybrid}. \rev{We tested multiple sets of cue weighting coefficients, including all configurations that HybridSORT uses across different datasets (e.g., appearance weight of 4.0 for DanceTrack, 1.3 for MOT17, and 4.6 for MOT20), and conducted a grid search over a neighborhood of these default values; the original DanceTrack configuration remained optimal.} The final assignment is obtained by finding a minimum cost assignment with the Hungarian algorithm on the final cost matrix.
As shown in Table~\ref{tab:heuristic-tracker}, our ALT layer significantly outperforms the hungarian baseline. This experiment also highlights the difficulty of designing effective heuristic trackers, which often require extensive hyperparameter tuning. For instance, HybridSORT uses different appearance weights across datasets - 4.0 for DanceTrack, 1.3 for MOT17, and 4.6 for MOT20 - and even varies weights based on detection confidence. In contrast, our learnable ALT module automatically learns to exploit the most relevant cues for different scenarios, eliminating the need for manual tuning.
\begin{table}[!ht]
\centering
\resizebox{\columnwidth}{!}{%
\begin{tabular}{l|ccc}
\hline\noalign{\smallskip}
Matching & HOTA & IDF1  & MOTA\\
\noalign{\smallskip}
\hline
\textit{\color{gray}{Online:}}  & & & \\
Hungarian  & 62.1 & 65.8  & \textbf{88.8} \\
ALT Layer (ours)            & \textbf{64.6}   & \textbf{69.0}    & 87.5    \\
\hline
\end{tabular}
}
\caption{Comparison of our \textit{online} ALT layer with the Hungarian algorithm on the DanceTrack val set.}
\label{tab:heuristic-tracker}
\end{table}

\PAR{Heuristic vs. Learnable Stitching. }
As explained in Section \ref{sec:autoreg-tracking}, SUSHI~\cite{Cetintas_2023_CVPR}, among other graph-based trakers \cite{braso_2020_CVPR, MPNTrackSeg, 10.1609/aaai.v38i3.27953, spam2024eccv}, uses Hungarian matching to stitch together the trajectories from a set of overlapping subclips and perform long-term tracking.
In contrast, our ALT layer learns to associate tracks across non-overlapping subclips in a fully data-driven manner, eliminating the need for explicit stitching and reducing redundant computations. To assess the effectiveness of this approach, we retrain SUSHI on DanceTrack using the same node/edge features and ReID model as NOOUGAT. As shown in Table~\ref{tab:heuristic-stitch}, NOOUGAT achieves a 3.2 IDF1 improvement, highlighting the superiority of learned associations over heuristic stitching.

\begin{table}[!ht]
\centering
\resizebox{\columnwidth}{!}{%
\begin{tabular}{l|ccc}
\hline\noalign{\smallskip}
Stitching & HOTA & IDF1  & MOTA\\
\noalign{\smallskip}
\hline
\textit{\color{gray}{Offline:}}  & & & \\
 SUSHI \cite{Cetintas_2023_CVPR}        & 62.1 & 62.4  & 88.8 \\
 SUSHI \cite{Cetintas_2023_CVPR}\footnote{}        & 67.9 & 72.0  & 88.9 \\
NOOUGAT (ours)             & \textbf{70.5} & \textbf{76.6}  & \textbf{89.0} \\
\hline
\end{tabular}
}
\caption{Comparison of NOOUGAT with our learnable ALT layer with the heuristic stitching in SUSHI on the DanceTrack val set. \rev{The first row shows original SUSHI performance; the second row shows our reproduction with matched features for fair comparison.}}
\label{tab:heuristic-stitch}
\end{table}

\footnotetext{Our SUSHI~\cite{Cetintas_2023_CVPR} reproduction with a ReID model trained on DanceTrack and our updated node and edge features}

\PAR{Association Cues.}
While heuristic methods explicitly define the relative importance of each cue, our model learns to infer the most relevant cues for each scenario. To better understand which cues contribute most to performance, we evaluate our model using different combinations of input features.
We categorize cues into three groups: Appearance (A), which includes ReID features; Motion (M), which combines motion prediction and our VDC feature; and Geometry (G), which uses only bounding box coordinates and temporal distance. As shown in Table~\ref{tab:cues-ablat}, \rev{interestingly, the appearance-only model performs \textit{worse} than the geometry-only model. We attribute this to the very high appearance similarities in DanceTrack, making appearance useful for disambiguating uncertain matches but unreliable in isolation. This aligns with prior work: QDTrack~\cite{qdtrack_conf}, an appearance-only tracker, performs worse than motion-only OC-SORT~\cite{cao2023observation} on DanceTrack. Conversely, DanceTrack's high frame rate makes geometry alone reasonably effective due to small inter-frame displacements.} Overall, NOOUGAT performs best when all cues are available, demonstrating that the model learns to leverage the most informative cues for different scenarios.

\begin{table}[!ht]
\centering
\resizebox{\columnwidth}{!}{%
\begin{tabular}{ccc|ccc}
\toprule
G & M & A & HOTA & IDF1 & MOTA \\
\midrule
\cmark & \xmark & \xmark  & 59.1  & 61.9 & 89.0 \\
\xmark & \xmark & \cmark  & 57.9  & 58.1 & 89.0 \\
\cmark & \cmark & \xmark  & 60.4  & 64.4 & 88.9 \\
\cmark & \cmark & \cmark  & 64.6  & 69.0 & 87.5 \\
\bottomrule
\end{tabular}
}
\caption{Ablation of our \textit{online} tracker on DanceTrack val set, using different tracking cues: Appearance (A), Motion and Velocity (M) and Geometry (G).}
\label{tab:cues-ablat}
\end{table}

\PAR{Training Parameters.}
Table~\ref{tab:alt-ablat} reports the impact of our VDC extension and data augmentation strategy (see Section~\ref{sec:unif-arch-train}) when training our \textit{offline} tracker. Notably, the inclusion of VDC yields a 1.4-point improvement in IDF1, underscoring its effectiveness even when combined with strong cues such as appearance and motion. 

\begin{table}[!ht]
\centering
\resizebox{\columnwidth}{!}{%
\begin{tabular}{l|ccc}
\hline\noalign{\smallskip}
Tracker & HOTA & IDF1  & MOTA\\
\noalign{\smallskip}
\hline
\textit{\color{gray}{Offline:}}  & & & \\
NOOUGAT                        & 68.4  &  73.5  & 89.0\\
\hspace{0.01cm} + VDC          & 69.5  &  74.9  & 89.0\\
\hspace{0.01cm} + Skip          & 69.8  &  75.2  & 89.0\\
\hspace{0.01cm} + Past tracks aug.          & \textbf{70.5}  &  \textbf{76.6}  & \textbf{89.0}\\
\hline
\end{tabular}
}
\caption{Ablations of our training parameters on the DanceTrack val set.}
\label{tab:alt-ablat}
\end{table}

\rev{
\PAR{Runtime Analysis.}
We analyze NOOUGAT's runtime characteristics to assess practical deployment viability.

\noindent\textit{Scalability with object count.} Table~\ref{tab:FPS-num-objects-main} reports runtime on MOT20-05, the densest sequence across all benchmarks (1,211 IDs, 751k ground-truth boxes, 3,315 frames). We sample increasingly larger subsets of IDs while retaining false positive detections for realistic conditions. With 100 IDs (54k detections---already exceeding any DanceTrack or SportsMOT sequence), NOOUGAT processes at 257 FPS. Even with all 1,154 IDs (613k detections), we maintain 74 FPS.

\begin{table}[!ht]
  \centering
  \begin{tabular}{c|c|c}
  \hline\noalign{\smallskip}
  Num. IDs & Num. Detections & FPS\\
  \noalign{\smallskip}
  \hline
  100 & 54,203 & 257.0 \\
  200 & 115,057 & 207.6 \\
  400 & 224,798 & 155.0 \\
  600 & 324,513 & 123.6 \\
  800 & 420,711 & 101.7 \\
  1000 & 530,166 & 83.1 \\
  1154 (All) & 612,988 & 74.0 \\
  \hline
  \end{tabular}
  \caption{Tracking runtime of NOOUGAT Offline on MOT20-05 for different numbers of IDs.}
  \label{tab:FPS-num-objects-main}
\end{table}

\noindent\textit{GPU memory.} On DanceTrack-val, NOOUGAT inference requires less than 6GB of VRAM in both online and offline ($T{=}256$) modes, making it suitable for consumer-grade hardware.

\noindent\textit{Model inference time.} We evaluate NOOUGAT's runtime on a single NVIDIA V100 GPU, with precomputed detections and ReID features. On the DanceTrack validation set, our \textit{online} model achieves an average runtime of 12 FPS, while the \textit{offline} model reaches 340 FPS. This discrepancy arises primarily from graph construction overhead, which accounts for approximately 48\% of the forward pass time in the \textit{online} setting, as a new graph is built at every frame. In contrast, the \textit{offline} model benefits from the GNN hierarchy, which significantly reduces the number of graphs constructed (e.g., only 8 graphs for a 256-frame window).
While our current implementation prioritizes correctness and modularity, we acknowledge that runtime efficiency---particularly in the \textit{online} mode---can be further optimized. Importantly, latency improves rapidly for $T{>}1$, as shown in Table~\ref{tab:FPS-oracle-incom-frames-main}: the oracle runtime already reaches 37 FPS at $T{=}4$ and 58 FPS at $T{=}8$. We believe NOOUGAT already offers substantial value for many applications in its present form.


\begin{table}[!ht]
  \centering
  \resizebox{\columnwidth}{!}{
  \begin{tabular}{c|c|c}
  \hline\noalign{\smallskip}
  Incom. Frames ($T$) & Hierarchy Layers & FPS\\
  \noalign{\smallskip}
  \hline
    1 & 0 &   18.6 \\
    4 & 2 &   37.3 \\
    8 & 3 &   57.8 \\
   16 & 4 &   94.2 \\
   64 & 6 &  251.1 \\
  256 & 8 &  695.0 \\
  \hline
  \end{tabular}}
  \caption{Oracle runtime on DanceTrack val set for different numbers of incoming frames $T$.}
  \label{tab:FPS-oracle-incom-frames-main}
\end{table}
}

\PAR{Towards an Application-Centric Tracker.}
\label{par:app-centric-track}
Finally, to assess the impact of temporal context on tracking performance, we ablate our key hyperparameter: the processing stride $T$. Recall that increasing $T$ allows the GNN hierarchy to jointly process a larger number of incoming frames, thereby providing more context and enabling richer temporal reasoning. We train multiple NOOUGAT configurations with $T$ ranging from 1 (\textit{online}) to 256 (our default \textit{offline} value). To isolate the effect of $T$, we disable the EMA of appearance features previously used with $T=1$ (see Section~\ref{sec:implem-details}).
As shown in Figure~\ref{fig:incom-frames-ablat}, increasing the number of incoming frames steadily improves performance.
To better understand the practical implications of this trend, we consider a range of real-world applications and their latency constraints, assuming a 30 FPS input stream. In this setting, each additional incoming frame adds approximately 33ms of latency budget.
While these constraints are often not rigid, we propose an overview of how different numbers of incoming frames $T$ align with application-specific requirements:
\begin{enumerate}
    \item Autonomous Driving  is latency-critical, and requires the perception stack to operate within 100ms \cite{10.1145/3296957.3173191}. This leaves little room for the tracker to await multiple incoming frames, thus limiting $T$ to 1 or 2.
    \item CCTV monitoring requires sufficient temporal resolution to detect security threats and incidents. Prior work \cite{10.1145/1459359.1459527} suggests that 8 FPS is adequate for theft detection, corresponding to a latency budget of ~125ms, or $T=3$–4.
    \item Telesurgery is optimal under 200ms, with 300ms being the upper bound for acceptable performance~\cite{telesurgery}, allowing for $T=6$–9.
    \item Aerial Vehicle Tracking often operates at low frame rates (1–2 FPS), as suggested by datasets such as~\cite{10.1007/978-3-031-73013-9_4, schmidt2012ais}. This permits significantly larger strides, like $T=15$–30.
    \item Offline Applications, such as visual effects, sports analytics and dataset annotation prioritize accuracy over latency. These can afford arbitrarily large $T$ values, as processing time is not a rigid constraint.
\end{enumerate}
This ablation highlights NOOUGAT’s versatility: by adjusting $T$, it can be configured to meet the latency and performance demands of a wide range of deployment scenarios, from real-time systems to offline analytics.

\rev{
\subsection{Generalization}
\label{sec:generalization}

To assess NOOUGAT's real-world applicability, we evaluate on additional benchmarks featuring diverse scenarios: BEE24 for challenging appearance similarity, VETRA for aerial vehicle tracking at extreme frame rates, and MOT17 with public detections for noisy detection conditions.

\PAR{BEE24.} BEE24~\cite{10851814} is a recent MOT benchmark for tracking bees, featuring complex motion patterns, heavy occlusions, challenging re-identification, and long sequences (up to 5,000 frames). As shown in Table~\ref{tab:BEE24-main}, NOOUGAT achieves +4.5 HOTA and +5.1 AssA over the next-best method TOPICTrack~\cite{10851814}, confirming our method's transferability to new domains without task-specific adaptations, and robustness to erratic motion and extremely similar appearances.

\begin{table}[!ht]
  \centering
  \resizebox{\columnwidth}{!}{%
  \begin{tabular}{l|cccc}
  \hline\noalign{\smallskip}
  Tracker & HOTA & IDF1 & AssA  & MOTA\\
  \noalign{\smallskip}
  \hline
  \textit{\color{gray}{Online Trackers:}}  & & & & \\
  TrackFormer \cite{meinhardt2021trackformer}                                                & 44.3 & 53.9 & 42.3 & 41.5 \\
  \rowcolor{salmon!15} ByteTrack \cite{zhang2022bytetrack}                             & 42.3 & 56.8 & 38.3 & 59.2 \\
  \rowcolor{salmon!15} OC-SORT \cite{cao2023observation}                               & 42.7 & 55.3 & 36.8 & 61.6 \\
  \rowcolor{salmon!15} TOPICTrack \cite{10851814}                           & 46.6 & 59.7 &   40.3  & 66.7 \\
  \rowcolor{salmon!15} \textbf{\modelname}  (ours)            & \textbf{51.1}   & \textbf{65.6}    &   \textbf{45.4}  & \textbf{72.9}    \\
  \hline
  \end{tabular}
  }
  \caption{Results on the BEE24 test set. Methods in the red block share the same detections.}
  \label{tab:BEE24-main}
\end{table}

\begin{table}[!ht]
    \centering
    \resizebox{\columnwidth}{!}{%
    \begin{tabular}{l|cccc}
    \hline\noalign{\smallskip}
    Tracker & HOTA$\uparrow$ & IDF1$\uparrow$ & MOTA$\uparrow$ & IDSW$\downarrow$\\
    \noalign{\smallskip}
    \hline
    \textit{\color{gray}{Online Trackers:}}  & & & & \\
    ByteTrack \cite{zhang2022bytetrack} & 36.4 & 17.8 & 13.6 & 17,328 \\
    BOT-SORT \cite{aharon2022botsortrobustassociationsmultipedestrian} & 50.8 & 48.2 & 18.5 & 6,886 \\
    Deep OC-SORT \cite{10222576} & 46.8 & 31.2 & 44.7 & 10,334 \\
    Deep OC-SORT\textsubscript{DIoU} \cite{10222576} & 39.5 & 32.1 & 32.6 & 13,068 \\
    \color{gray}{Deep SR-SORT} \cite{10.1007/978-3-031-73013-9_4} & \color{gray}{82.2} & \color{gray}{90.3} & \color{gray}{88.5} & \color{gray}{792} \\
    \modelname (ours) & 68.0 & 73.4 & 61.5 & 1,855 \\
    \hline
    \end{tabular}
    }
    \caption{Results on the VETRA test set. All methods share the same detections. Methods in gray leverage external priors such as average detection size and vehicle speed.}
    \label{tab:VETRA-main}
\end{table}

\begin{table}[!ht]
  \centering
  \resizebox{\columnwidth}{!}{%
  \begin{tabular}{l|cccc}
  \hline\noalign{\smallskip}
  Method & HOTA$\uparrow$ & IDF1$\uparrow$ & AssA$\uparrow$ & MOTA$\uparrow$\\
  \noalign{\smallskip}
  \hline
  \textit{\color{gray}{Online Trackers:}} & & & & \\
  Tracktor~\cite{tracktor_2019_ICCV} & 44.8 & 55.1 & 45.1 & 56.3 \\
  GMT \cite{gmt} & 51.2 & 65.9 & 55.1 & 60.2 \\
  UTM \cite{10204724} & 52.5 & 65.1 & 53.2 & 63.5 \\
  OC-SORT \cite{cao2023observation} & 52.4 & 65.1 & \textbf{57.6} & 58.2 \\
  \textbf{\modelname} (ours) & \textbf{52.7} & \textbf{67.9} & 56.0 & \textbf{61.7} \\
  \hline
  \textit{\color{gray}{Offline Trackers:}} & & & & \\
  MPNTrack \cite{braso_2020_CVPR} & 49.0 & 61.7 & 51.1 & 58.8 \\
  Lif\_T \cite{lift} & 51.3 & 65.6 & 54.7 & 60.5 \\
  ApLift \cite{aplift} & 51.1 & 65.6 & 53.5 & 60.5 \\
  LPC\_MOT \cite{lpc} & 51.5 & 66.8 & 56.0 & 59.0 \\
  SUSHI \cite{Cetintas_2023_CVPR} & 54.6 & 71.5 & 59.5 & \textbf{62.0} \\

  \textbf{\modelname} (ours) & \textbf{54.9} & \textbf{72.1} & \textbf{60.2} & 61.5 \\
  \hline
  \end{tabular}
  }
  \caption{Results on the MOT17 test set with public detections. All methods use detections refined by Tracktor~\cite{tracktor_2019_ICCV}.}
  \label{tab:mot17-public-main}
  \end{table}
}

\PAR{VETRA.} VETRA~\cite{10.1007/978-3-031-73013-9_4} is a dataset for vehicle tracking from aerial imagery under extreme conditions: 1 FPS capture rate, 3 vehicle classes, and only 308 training images---17$\times$ fewer than MOT17 and 136$\times$ fewer than DanceTrack. To support VETRA's multi-class setup, we add a node feature indicating each detection's predicted class and a boolean edge feature for same-class detection pairs.

As shown in Table~\ref{tab:VETRA-main}, NOOUGAT outperforms ByteTrack and Deep OC-SORT by large margins. These methods struggle with VETRA's 1 FPS capture rate, which causes large inter-frame displacements. NOOUGAT achieves +17.2 HOTA and +25.2 IDF1 over BOT-SORT, ranking second only to Deep SR-SORT~\cite{10.1007/978-3-031-73013-9_4}---a method specifically crafted for this dataset that leverages external priors such as average detection size and vehicle speed. NOOUGAT learns these patterns in a data-driven manner despite the extreme data scarcity.

\PAR{MOT17 Public.} We train NOOUGAT on MOT17 public detections, refined by Tracktor~\cite{tracktor_2019_ICCV}, to evaluate performance under noisy detection conditions. As shown in Table~\ref{tab:mot17-public-main}, NOOUGAT performs favorably compared to other trackers, although it does so by a lesser margin than in the private detection setup. We hypothesize that noisy detections make it challenging for ALT to learn associations. However, recent progress in object detection \cite{yolox2021, rf-detr, huang2025deimv2} has made obtaining high-quality detections more straightforward on custom datasets.

\subsection{Qualitative Results}

\begin{figure*}[htbp]
  \centering
  \begin{subfigure}[b]{0.49\textwidth}
    \includegraphics[width=\linewidth]{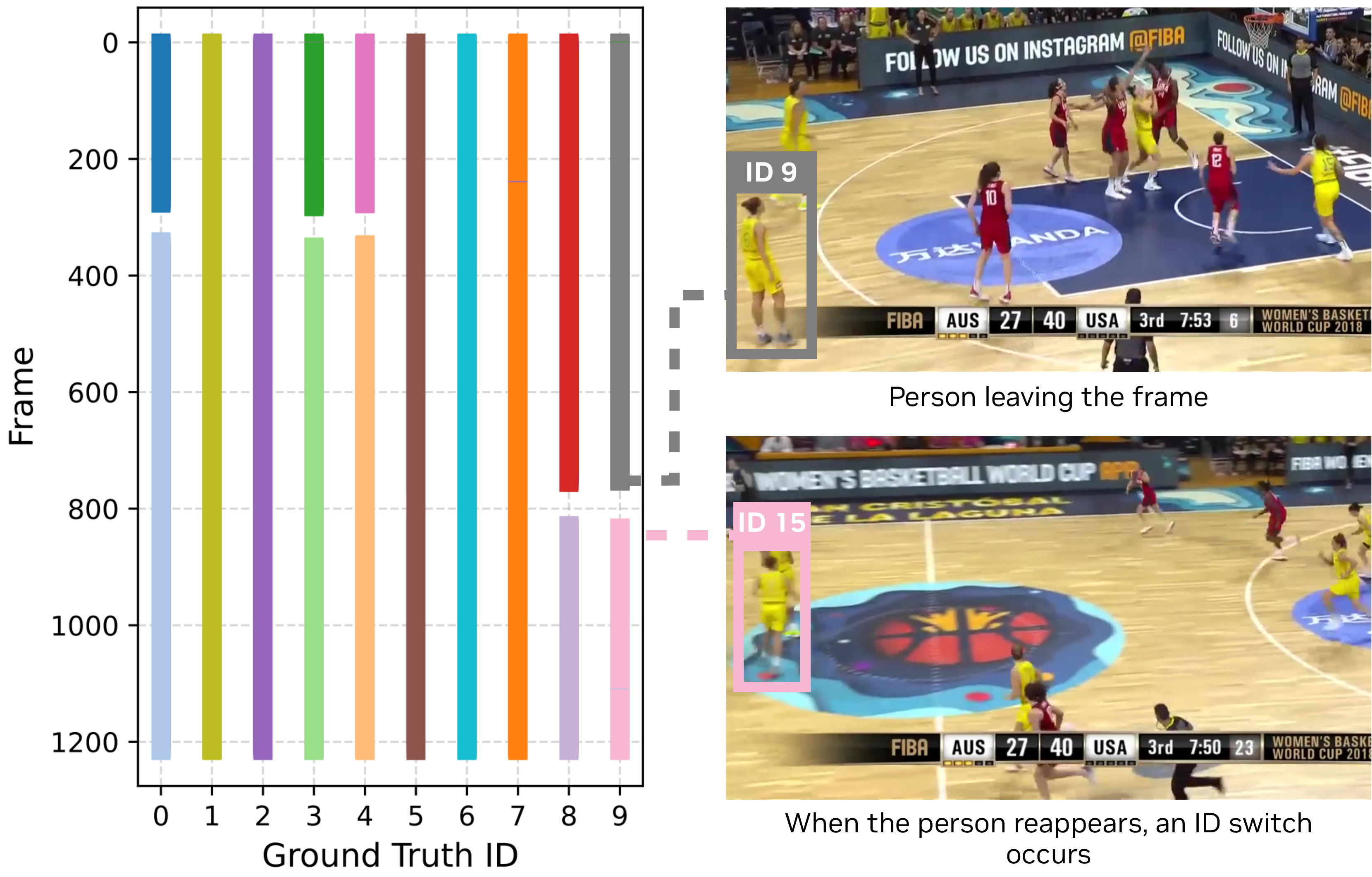}
    \caption{DiffMOT~\cite{lv2024diffmot}}
    \label{fig:qual-diffmot1}
  \end{subfigure}
  \hfill
  \begin{subfigure}[b]{0.49\textwidth}
    \includegraphics[width=\linewidth]{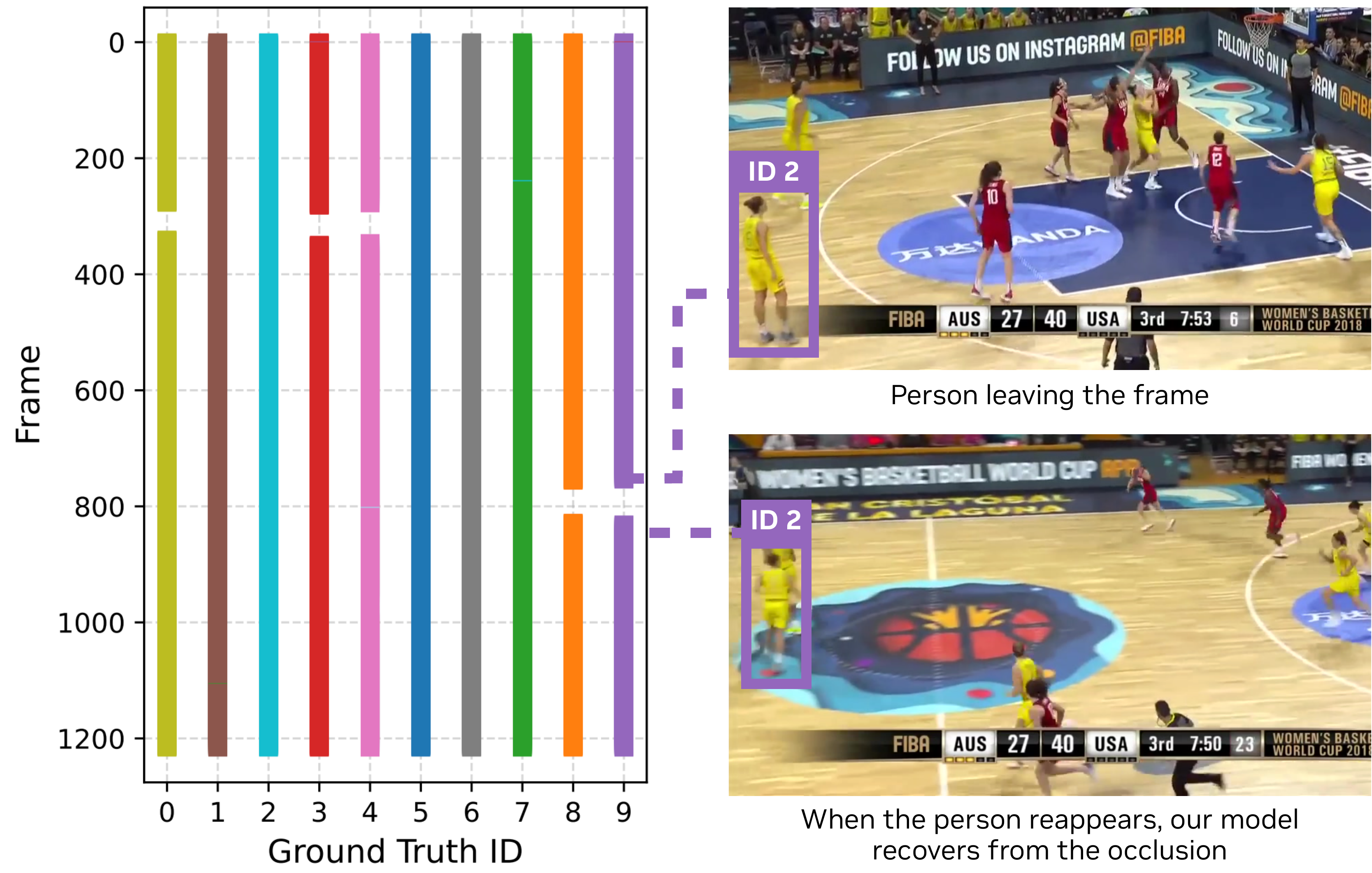}
    \caption{NOOUGAT Online (ours)}
    \label{fig:qual-noougat1}
  \end{subfigure}

  \setcounter{subfigure}{0}

  \begin{subfigure}[b]{0.49\textwidth}
    \includegraphics[width=\linewidth]{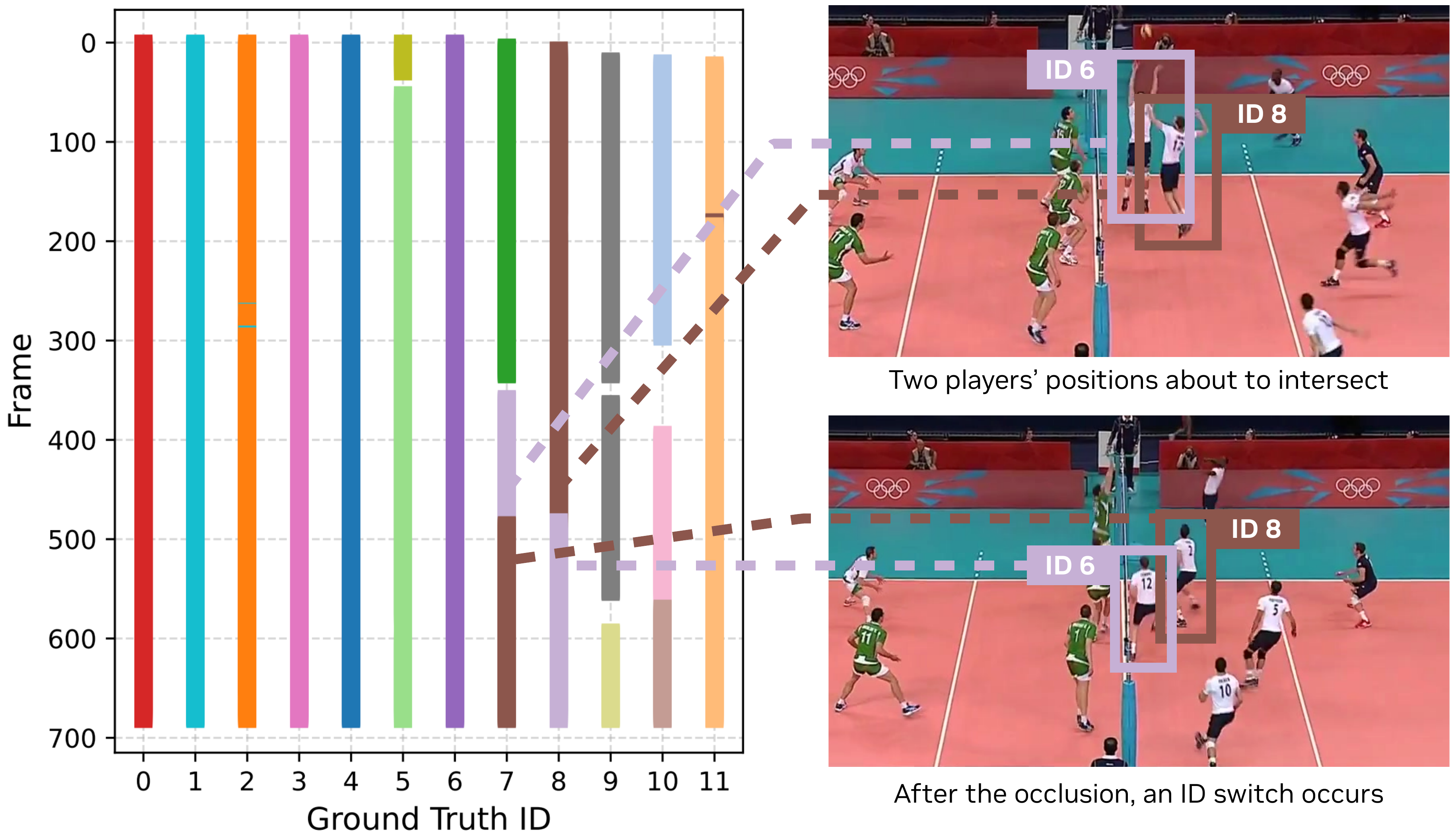}
    \caption{DiffMOT~\cite{lv2024diffmot}}
    \label{fig:qual-diffmot2}
  \end{subfigure}
  \hfill
  \begin{subfigure}[b]{0.49\textwidth}
    \includegraphics[width=\linewidth]{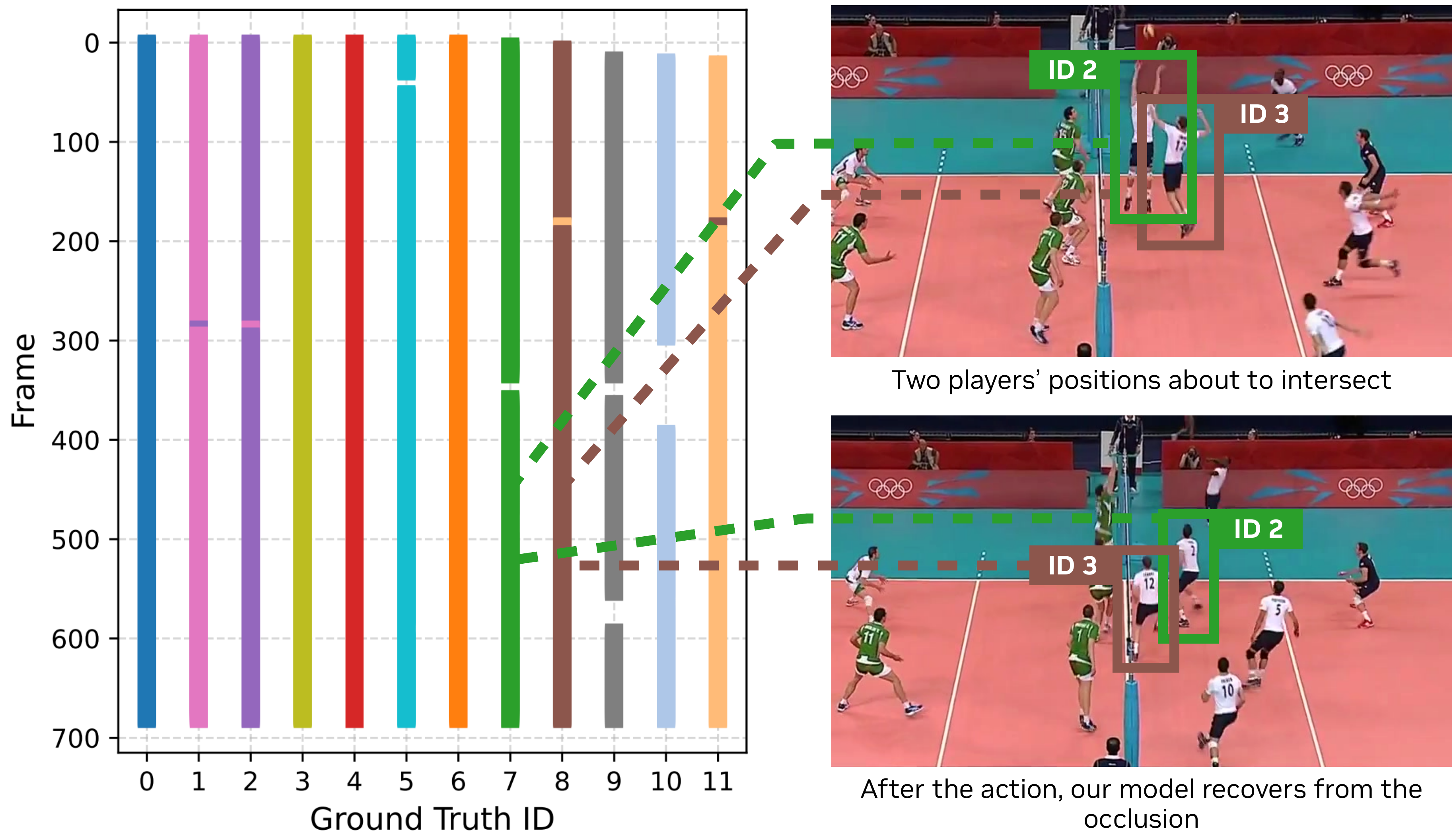}
    \caption{NOOUGAT Online (ours)}
    \label{fig:qual-noougat2}
  \end{subfigure}

  \setcounter{subfigure}{0}

  \begin{subfigure}[b]{0.49\textwidth}
    \includegraphics[width=\linewidth]{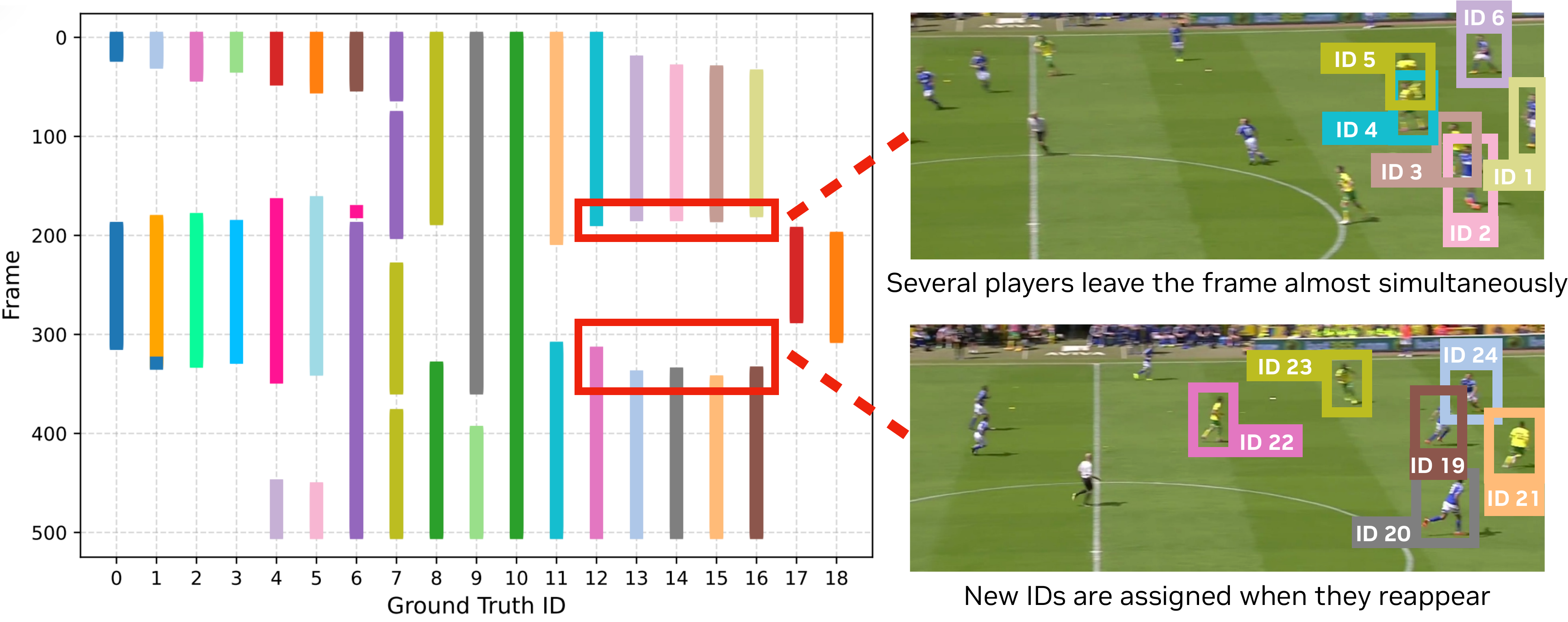}
    \caption{DiffMOT~\cite{lv2024diffmot}}
    \label{fig:qual-diffmot3}
  \end{subfigure}
  \hfill
  \begin{subfigure}[b]{0.49\textwidth}
    \includegraphics[width=\linewidth]{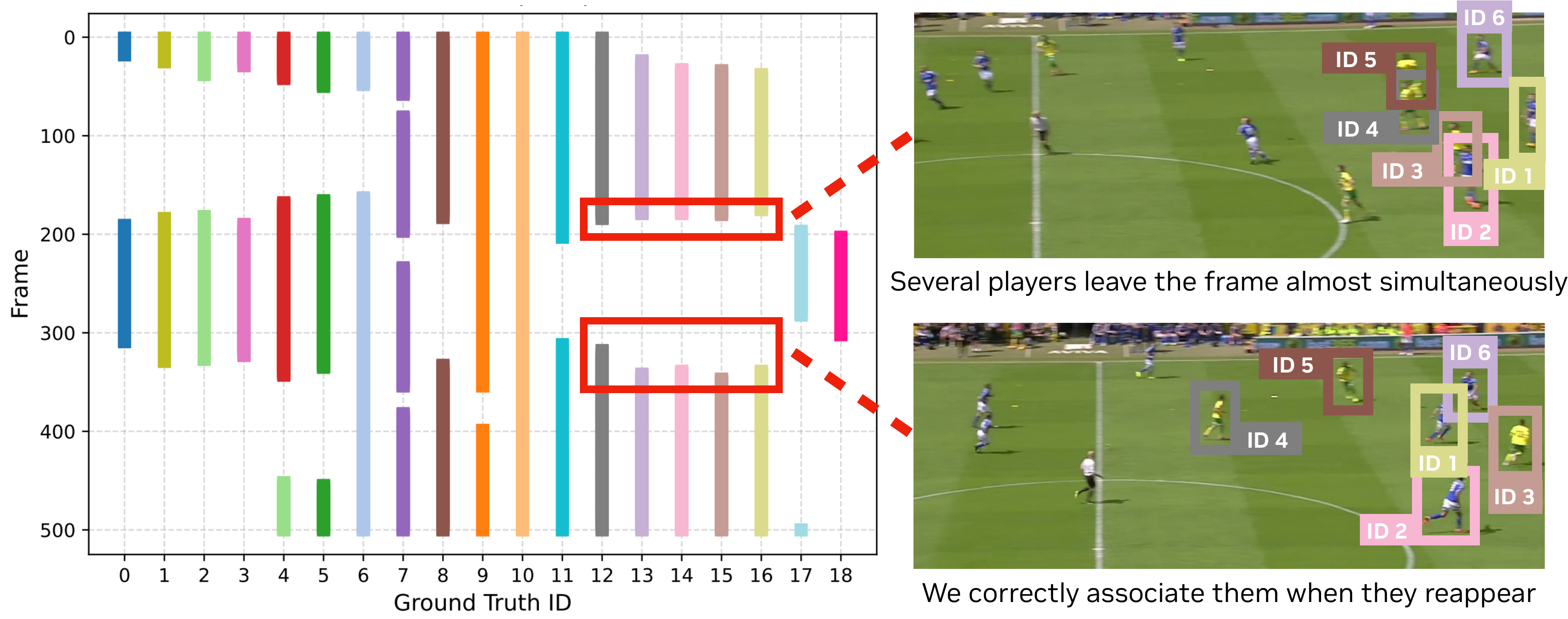}
    \caption{NOOUGAT Online (ours)}
    \label{fig:qual-noougat3}
  \end{subfigure}

  \caption{Qualitative comparisons between DiffMOT~\cite{lv2024diffmot} and NOOUGAT (ours) on selected SportsMOT validation sequences. For each ground truth ID, we visualize the predicted ID over time using a unique color. NOOUGAT exhibits better robustness to occlusions and complex player interactions.}
  \label{fig:qualitative_comparison}
\end{figure*}

Figure~\ref{fig:qualitative_comparison} provides qualitative comparisons between NOOUGAT and the next best tracker DiffMOT~\cite{lv2024diffmot} on selected sequences from the SportsMOT validation set. We plot the predicted IDs over time for each ground truth ID using distinct colors. The results demonstrate NOOUGAT’s ability to recover from occlusions and capture complex player interactions. We provide an additional qualitative comparison between NOOUGAT Offline and SUSHI~\cite{Cetintas_2023_CVPR} on selected DanceTrack validation sequences in Figure~\ref{fig:sushi_comparison_main}. NOOUGAT exhibits better robustness to occlusions and fewer track fragmentations.

\rev{
\begin{figure*}[htbp]
  \centering
  \includegraphics[width=\textwidth, height=0.35\textheight, keepaspectratio]{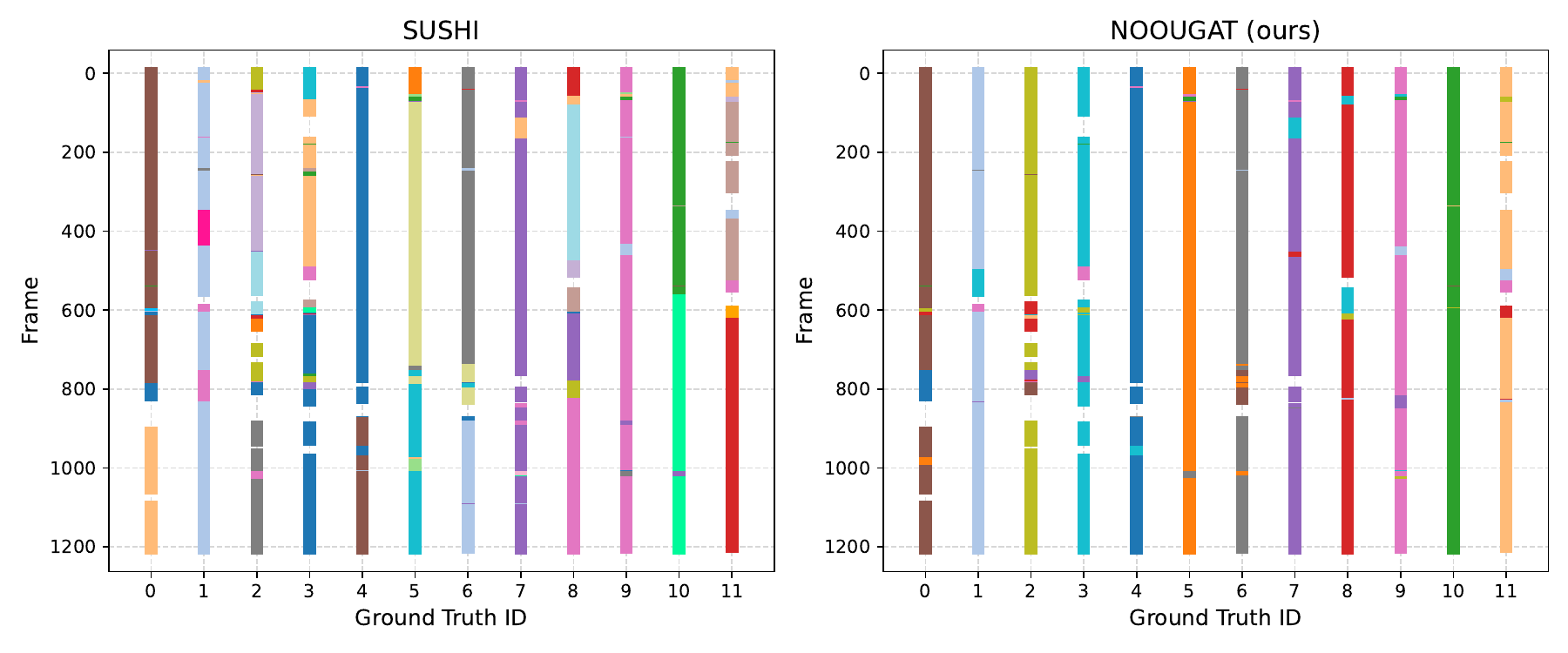}
  
  \vspace{6pt}
  
  \includegraphics[width=\textwidth, height=0.35\textheight, keepaspectratio]{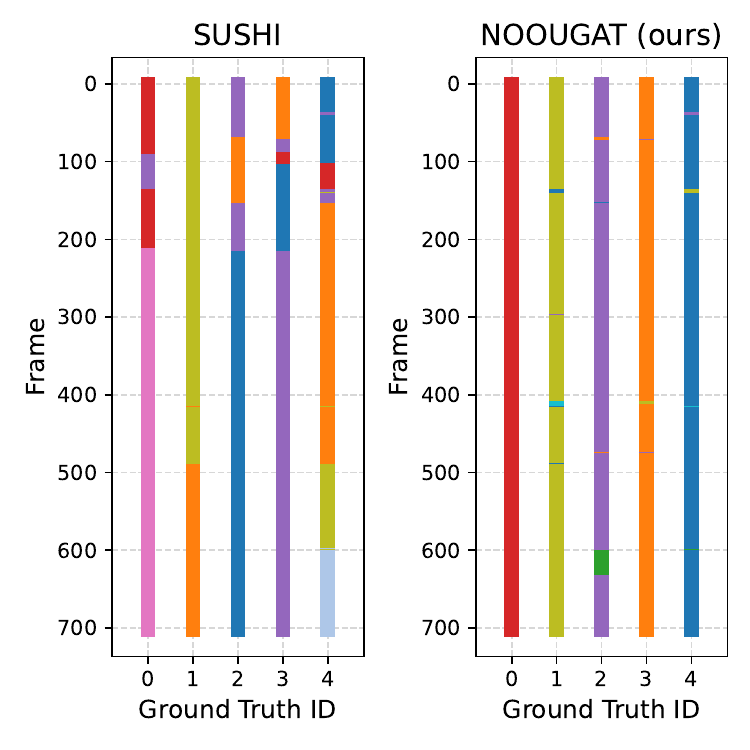}
  
  \caption{Qualitative comparison between SUSHI~\cite{Cetintas_2023_CVPR} and NOOUGAT Offline (ours) on selected DanceTrack validation sequences. NOOUGAT exhibits better robustness to occlusions and fewer track fragmentations.}
  \label{fig:sushi_comparison_main}
\end{figure*}
}

%

\subsection{Data Availability Statement}
The datasets used in this work are publicly available and can be accessed through the following links:
\begin{itemize}
    \item MOT17 \cite{dendorfer2020motchallengebenchmarksinglecameramultiple}: \href{https://motchallenge.net/data/MOT17}{motchallenge.net}
    \item MOT20 \cite{dendorfer2020motchallengebenchmarksinglecameramultiple}: \href{https://motchallenge.net/data/MOT20}{motchallenge.net}
    \item DanceTrack \cite{sun2022dance}: \href{https://dancetrack.github.io/}{dancetrack.github.io}
    \item SportsMOT \cite{cui2023sportsmot}: \href{https://github.com/MCG-NJU/SportsMOT}{github.com/MCG-NJU/SportsMOT}
\end{itemize}
 
All datasets are used under their respective licenses and terms of use. No new datasets were generated during this study.

\section{Conclusion}

In this work, we introduced NOOUGAT, the first tracker designed to flexibly adapt to a wide range of application constraints and deployment scenarios. Our experiments showed consistent improvements over existing \textit{online} and \textit{offline} methods, and our ablation studies highlighted the advantages of learned associations over heuristic matching and stitching - particularly in recovering from long-term occlusions. We hope this work will inspire the community to rethink the traditional separation between online and offline tracking and encourage a shift toward more application-oriented tracking approaches.

\newpage

\newpage
\bibliography{sn-bibliography}

\end{document}